\documentclass[runningheads,a4paper]{llncs}
\makeatletter
\@twosidefalse
\@mparswitchfalse
\makeatother

\usepackage{graphicx}
\usepackage[utf8]{inputenc}
\usepackage{newtxtext}
\usepackage{newtxmath}
\usepackage{newtxtt}
\usepackage{fontawesome}
\usepackage{courier}
\usepackage{booktabs}
\usepackage{tabularx, booktabs}
\newcolumntype{Y}{>{\centering\arraybackslash}X}

\usepackage{scalerel}
\usepackage{tikz}
\usetikzlibrary{svg.path}

\definecolor{orcidlogocol}{HTML}{A6CE39}
\tikzset{
  orcidlogo/.pic={
    \fill[orcidlogocol] svg{M256,128c0,70.7-57.3,128-128,128C57.3,256,0,198.7,0,128C0,57.3,57.3,0,128,0C198.7,0,256,57.3,256,128z};
    \fill[white] svg{M86.3,186.2H70.9V79.1h15.4v48.4V186.2z}
                 svg{M108.9,79.1h41.6c39.6,0,57,28.3,57,53.6c0,27.5-21.5,53.6-56.8,53.6h-41.8V79.1z M124.3,172.4h24.5c34.9,0,42.9-26.5,42.9-39.7c0-21.5-13.7-39.7-43.7-39.7h-23.7V172.4z}
                 svg{M88.7,56.8c0,5.5-4.5,10.1-10.1,10.1c-5.6,0-10.1-4.6-10.1-10.1c0-5.6,4.5-10.1,10.1-10.1C84.2,46.7,88.7,51.3,88.7,56.8z};
  }
}

\newcommand\orcidicon[1]{\href{https://orcid.org/#1}{\mbox{\scalerel*{
\begin{tikzpicture}[yscale=-1,transform shape]
\pic{orcidlogo};
\end{tikzpicture}
}{|}}}}

\usepackage{hyperref} 

\begin{document}
\title{Quick, Stat!: A Statistical Analysis of the Quick, Draw! Dataset}

\author{Raul Fernandez-Fernandez \orcidID{\orcidicon{0000-0002-6279-5553}}
$^{(\textsc{\faEnvelopeO})}$ \and
Juan G. Victores \orcidID{\orcidicon{0000-0002-3080-3467}} \and
David Estevez \orcidID{\orcidicon{0000-0002-8791-6649}} \and
Carlos Balaguer \orcidID{\orcidicon{0000-0003-4864-4625}}
}

\institute{All of the authors are members of the Robotics Lab research group within the Department of Systems Engineering and Automation, Universidad Carlos III de Madrid (UC3M).\\ \email{rauferna@ing.uc3m.es}}
\maketitle              
\begin{abstract}
\vspace*{-3.1em} 
The Quick, Draw! Dataset is a Google dataset with a collection of 50 million drawings, divided in 345 categories, collected from the users of the game Quick, Draw!\footnote{https://quickdraw.withgoogle.com/}. 
In contrast with most of the existing image datasets, in the Quick, Draw! Dataset, drawings are stored as time series of pencil positions instead of a bitmap matrix composed by pixels. This aspect makes this dataset the largest doodle dataset available at the time.
The Quick, Draw! Dataset is presented as a great opportunity to researchers for developing and studying machine learning techniques. 
Due to the size of this dataset and the nature of its source, there is a scarce of information about the quality of the drawings contained.
In this paper a statistical analysis of three of the classes contained in the Quick, Draw! Dataset is depicted: \emph{mountain}, \emph{book} and \emph{whale}. The goal is to give to the reader a first impression of the data collected in this dataset. 
For the analysis of the quality of the drawings a Classification Neural Network was trained to obtain a classification score. Using this classification score and the parameters provided by the dataset, a statistical analysis of the quality and nature of the drawings contained in this dataset is provided.
\end{abstract}
\section{Introduction}
\label{intro}

Since the introduction of Deep Learning techniques there has been a tendency to increase the number of layers in order to increase the performance of these architectures \cite{Zhang2018,Szegedy2015,Greff}. As a result of this tendency, in the present, it is common to find Deep Neural Network architectures with hundreds of layers. 
These architectures require very large amounts of training data in order to obtain useful results.
To address this problem, some techniques have been proposed to deal with the problem of lacking sufficient amounts of training data \cite{Perez2017} by generating new data from the limited available data. However, having a rich and diverse dataset is still a much more effective way to correctly train a Neural Network (NN) \cite{Deng2009,Xiao2017,Krizhevsky}. The Quick, Draw! Dataset \cite{quick-draw} is presented as the largest sketch dataset available at the time. The Quick, Draw! Dataset collects the sketches of the players of the Quick, Draw! open online game. In this game, the players are given a category they have to sketch in a maximum time limit of 20 seconds. At the same time, a classification Neural Network attempts to guess the category that the player is drawing. The player wins if the NN correctly guesses the category of what the player was prompted to sketch within the time limit. 

What is more interesting about this dataset is that it does not store the sketch performed by the player as a bitmap matrix composed by pixels. Instead, the sketch is saved as time series composed by pencil positions (x,y). Therefore, in addition to the spatial information that can be contained in an image, this dataset also contains temporal information of the sketch. As additional information, with each sketch, the dataset also include the category the player was asked to sketch, the timestamp when the sketch was created, the country where the player was located, and a unique sketch identifier. 

\begin{figure}[htpb]
  \centering
  \includegraphics[width=0.15\textwidth]{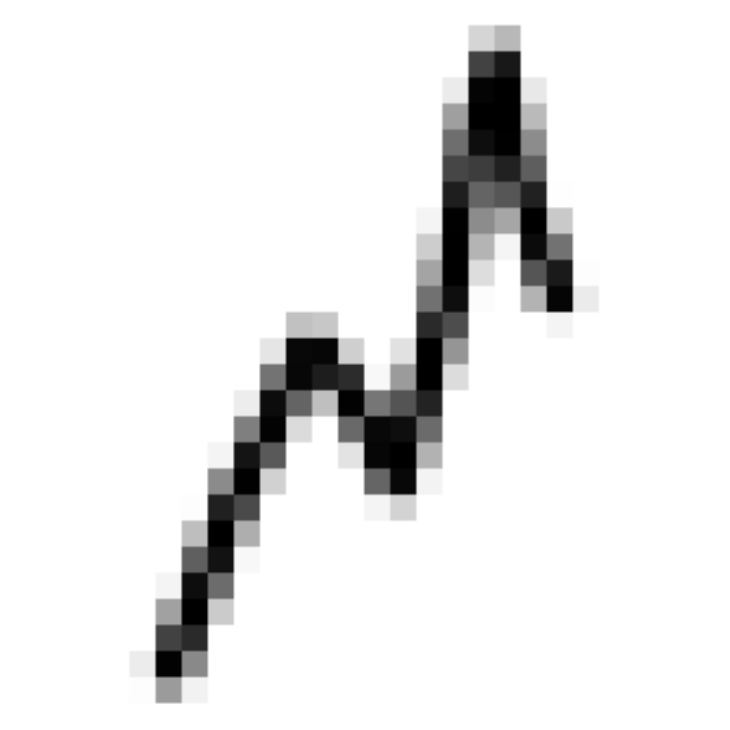}
  \includegraphics[width=0.15\textwidth]{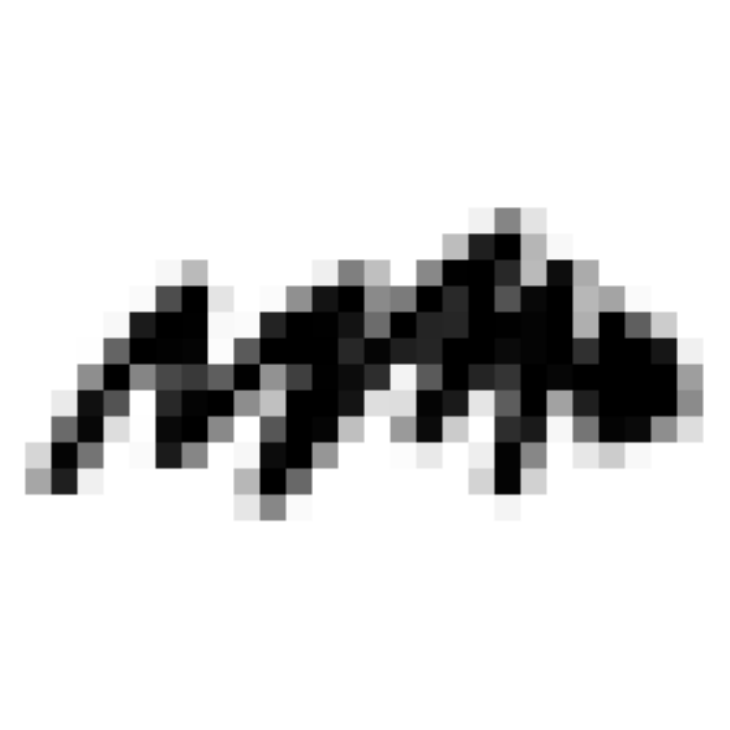}
  \includegraphics[width=0.15\textwidth]{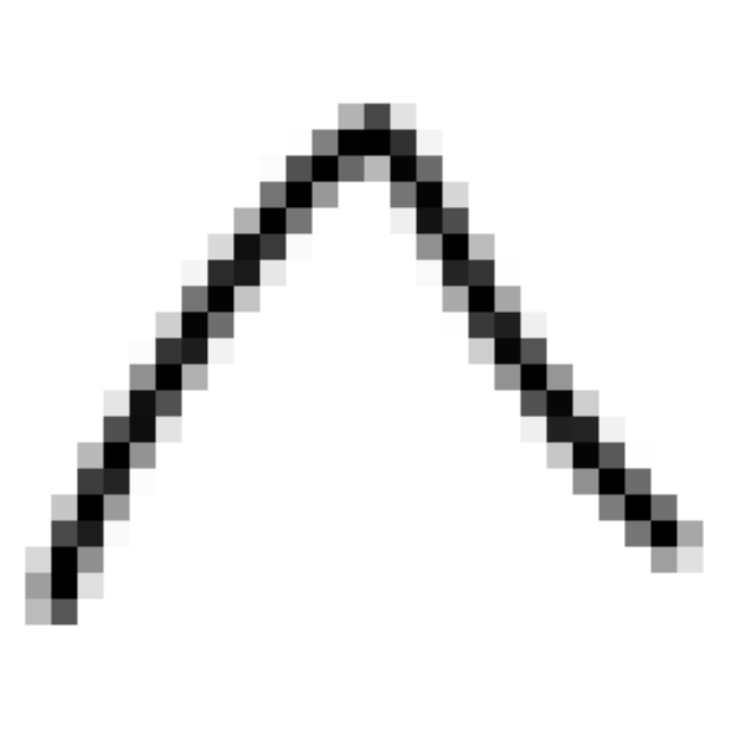}
  \includegraphics[width=0.15\textwidth]{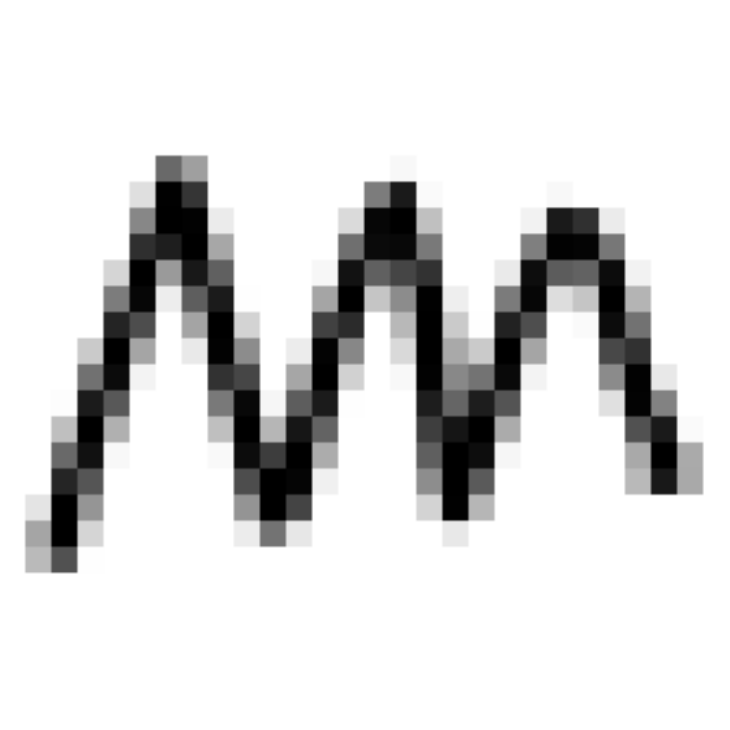}
  \includegraphics[width=0.15\textwidth]{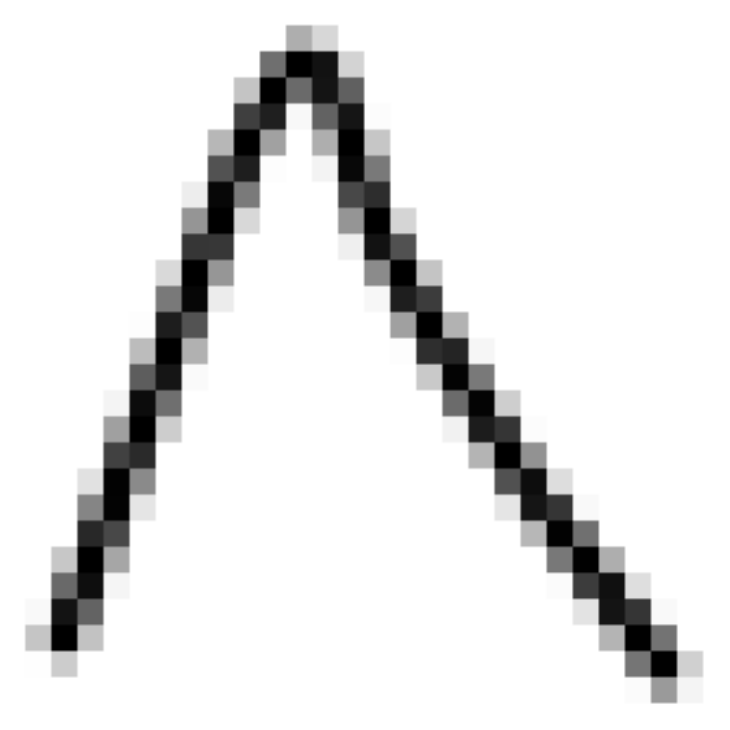}
  \includegraphics[width=0.15\textwidth]{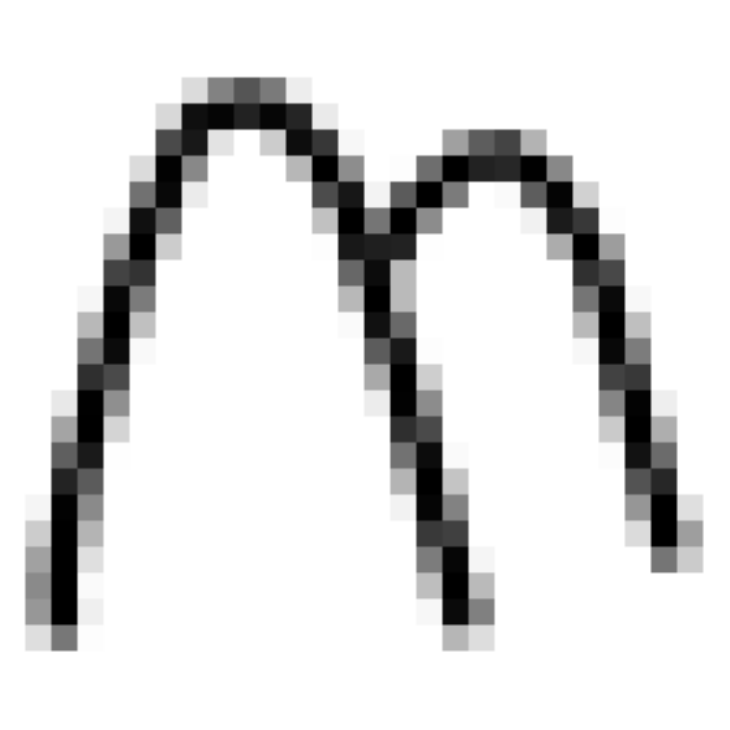}
  \\
  
  \includegraphics[width=0.15\textwidth]{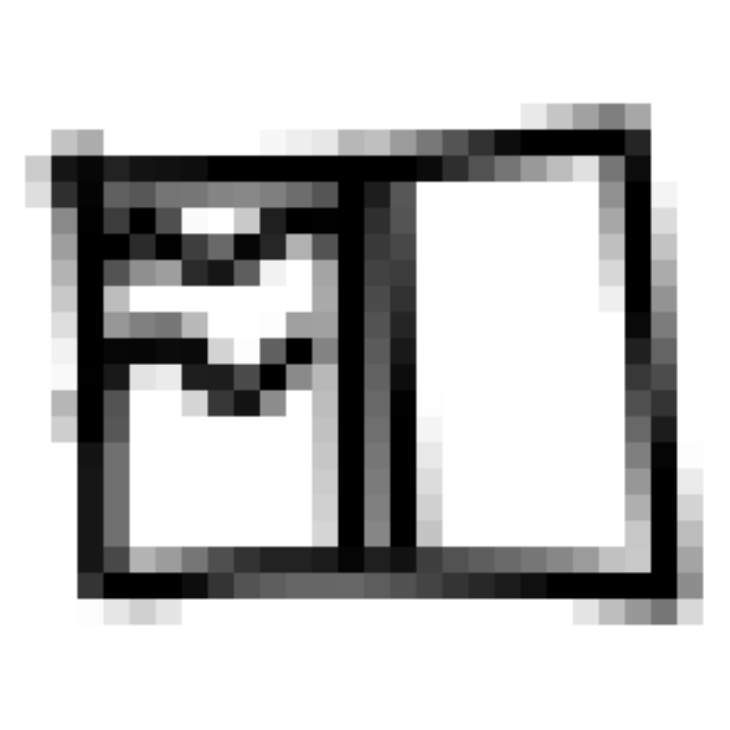}
  \includegraphics[width=0.15\textwidth]{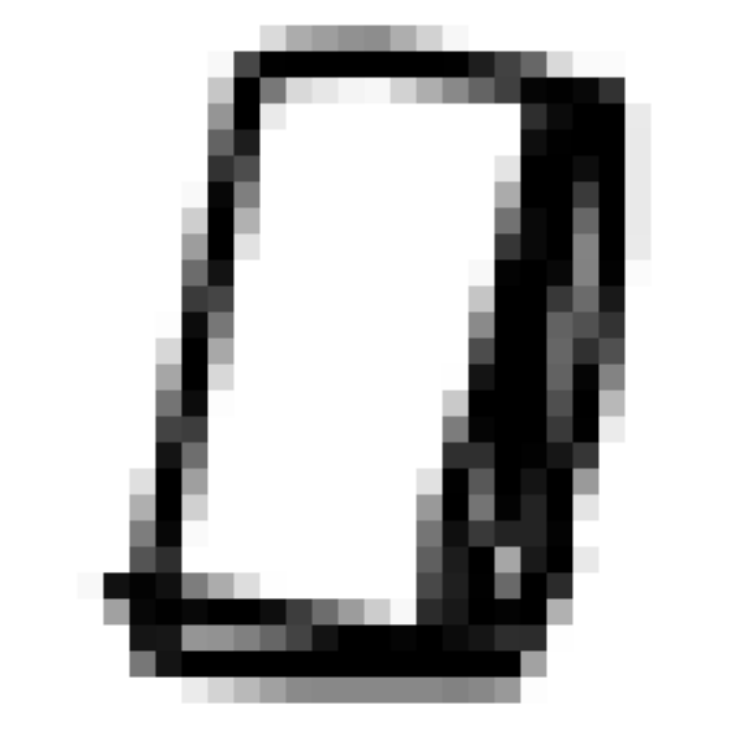}
  \includegraphics[width=0.15\textwidth]{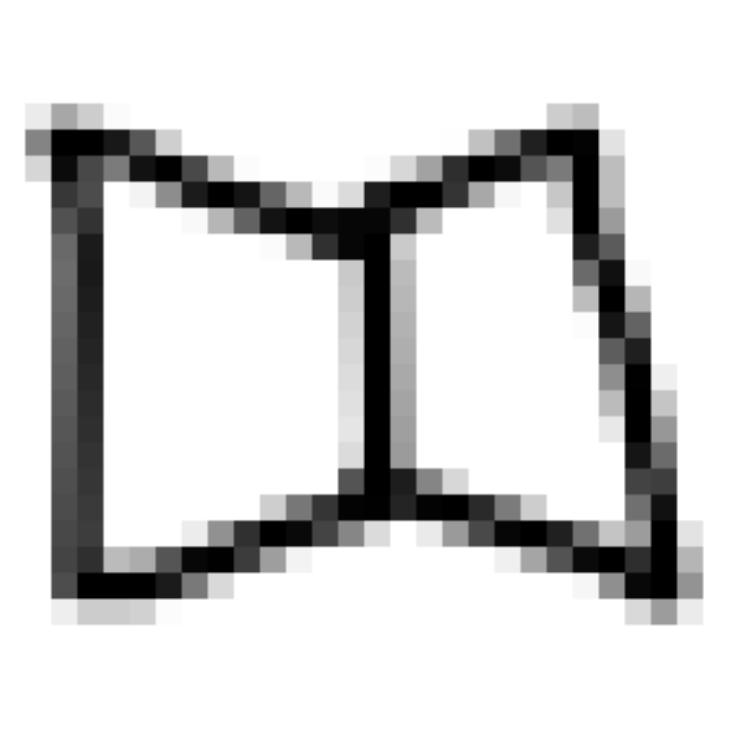}
  \includegraphics[width=0.15\textwidth]{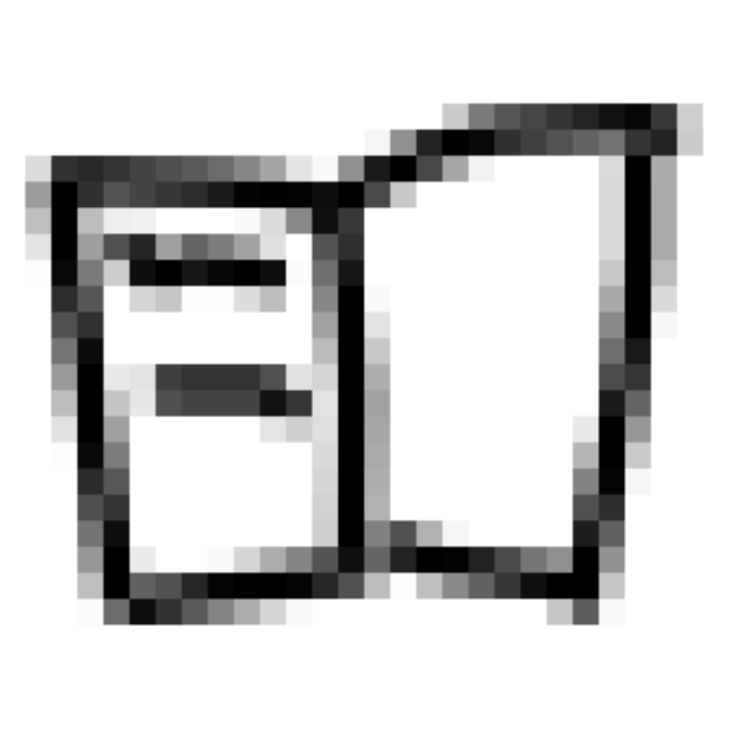}
  \includegraphics[width=0.15\textwidth]{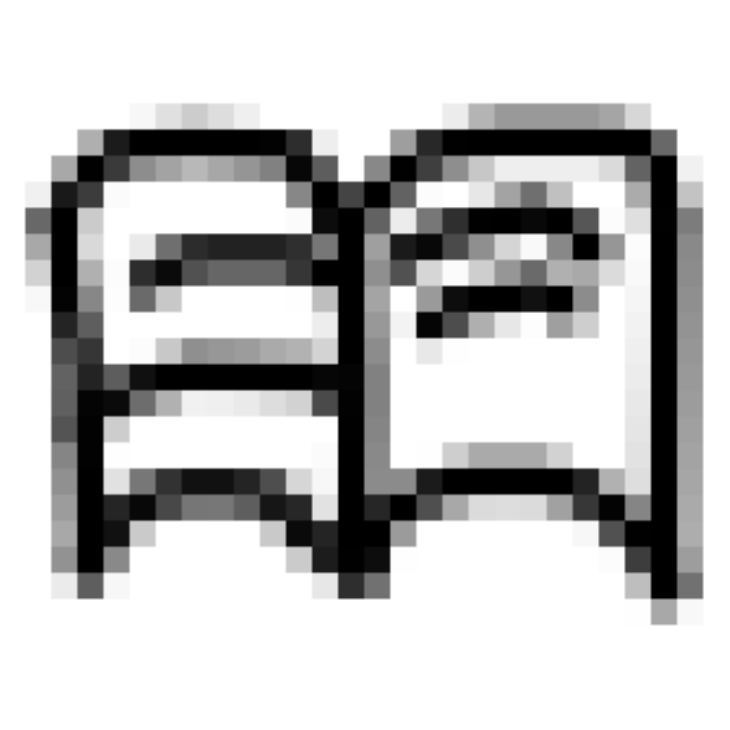}
  \includegraphics[width=0.15\textwidth]{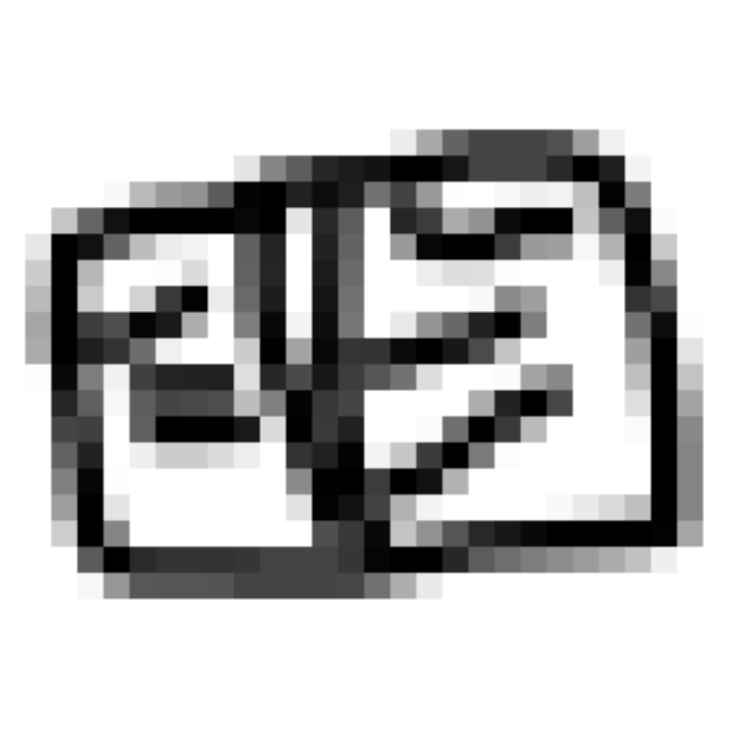}
  \\
  
  \includegraphics[width=0.15\textwidth]{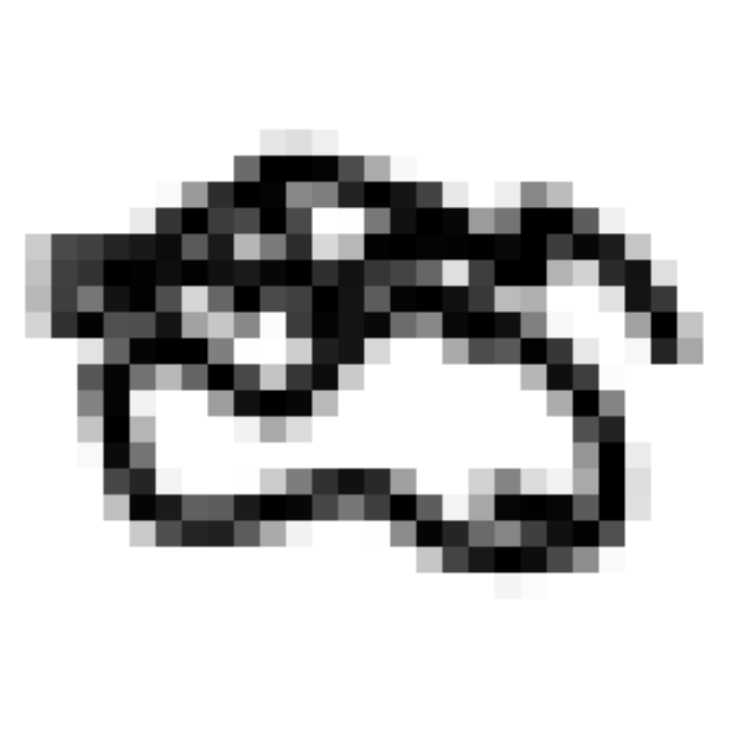}
  \includegraphics[width=0.15\textwidth]{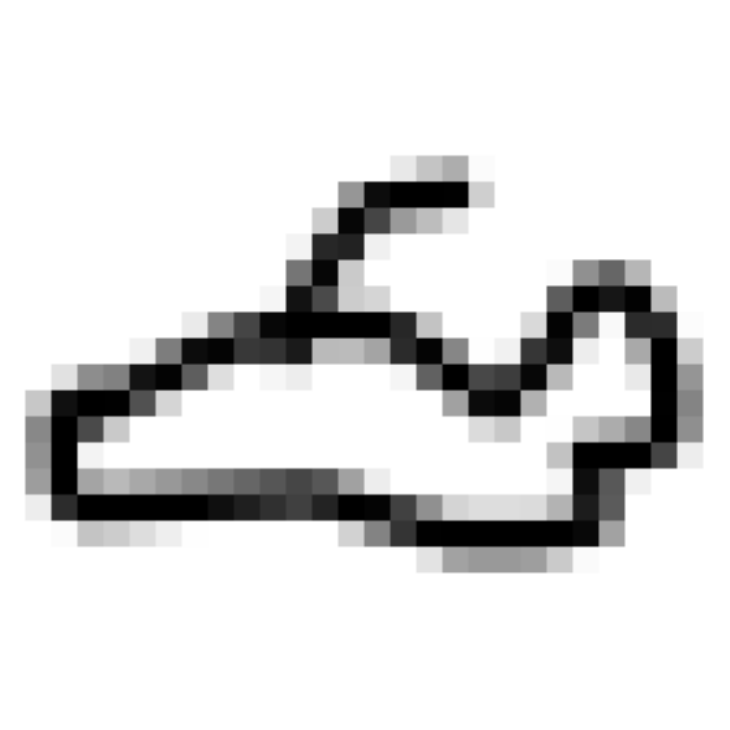}
  \includegraphics[width=0.15\textwidth]{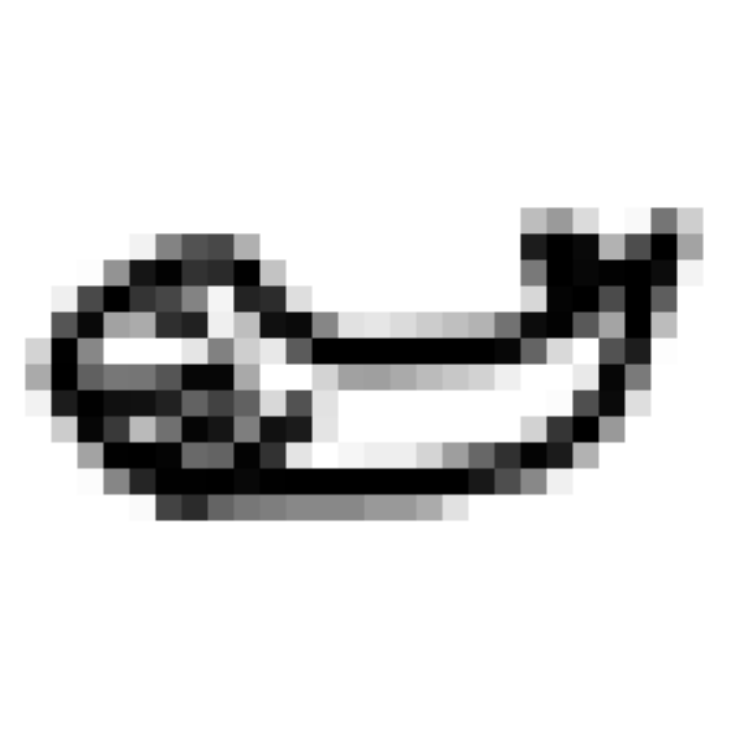}
  \includegraphics[width=0.15\textwidth]{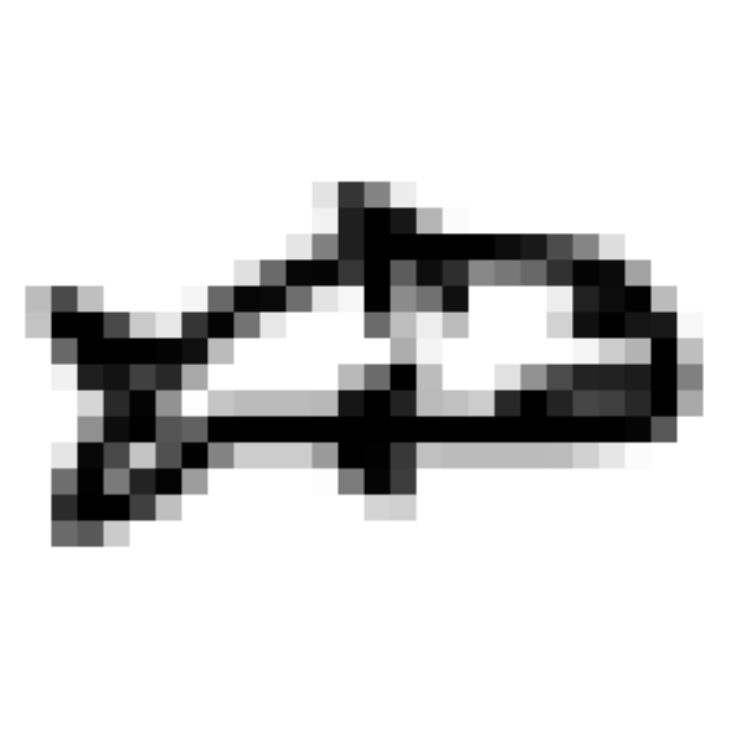}
  \includegraphics[width=0.15\textwidth]{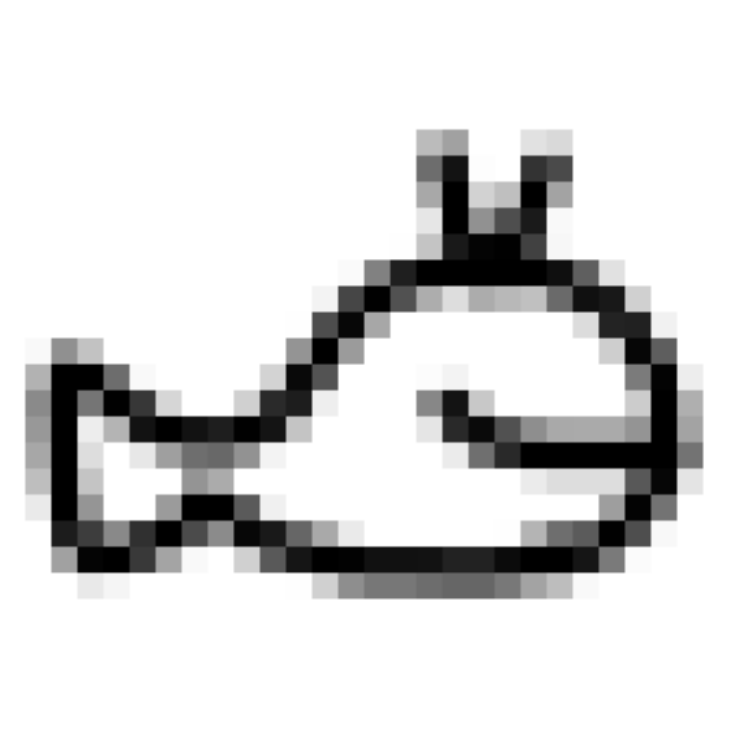}
  \includegraphics[width=0.15\textwidth]{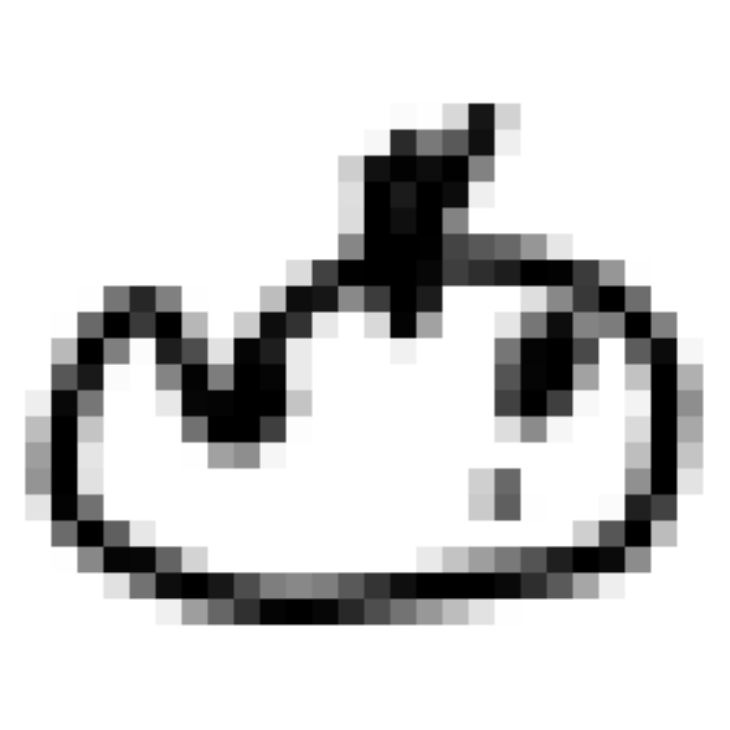}
  \\
  
  \includegraphics[width=0.15\textwidth]{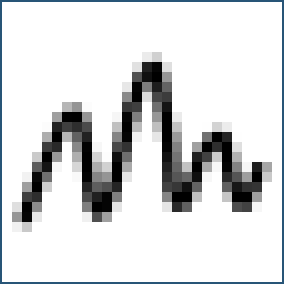}
  \includegraphics[width=0.15\textwidth]{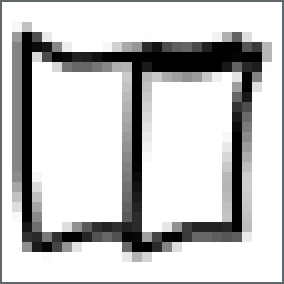}
  \includegraphics[width=0.15\textwidth]{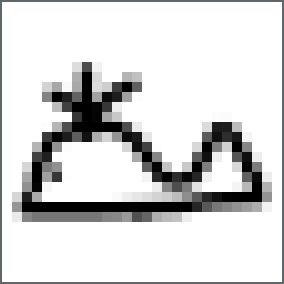}
  \\
  
  \hspace*{-0.1cm}
  \includegraphics[width=0.15\textwidth]{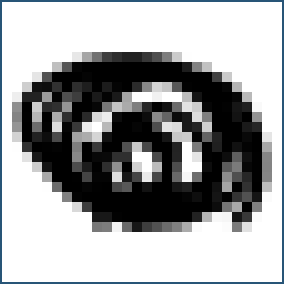}
  \includegraphics[width=0.15\textwidth]{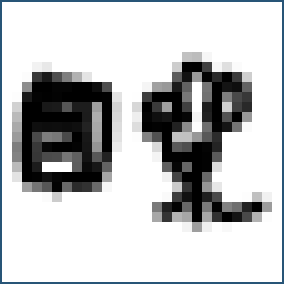}
  \includegraphics[width=0.15\textwidth]{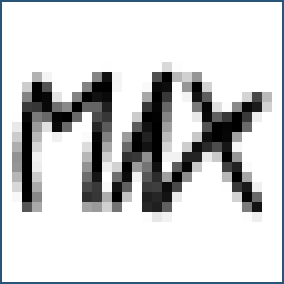}

  \caption{First three rows: Random examples extracted from the Quick, Draw! Dataset of the three categories studied in this paper. From top to bottom: \emph{mountain}, \emph{book}, \emph{whale}. Last two rows: The first row depicts good sketches with high classification scores. The second row depicts examples of bad sketches with low classification scores. From left to right: \emph{mountain}, \emph{book}, \emph{whale}.}
  \label{fig:3class}
\end{figure}

In this paper, a statistical analysis of three of the categories presented in this dataset: \emph{mountain}, \emph{book} and \emph{whale}, is performed. These three categories were selected to be different enough in terms of complexity to be considered valid representatives of the wide spectrum of sketches in the Quick, Draw! Dataset. The \emph{mountain} category is considered an example of a low complexity sketch; the \emph{whale} category is a representative of a more complex spectrum of sketches; while the \emph{book} category can be considered an average complexity sketch, see Fig. \ref{fig:3class} for reference. From each of the sketches presented in each of the categories, three parameters are extracted: the classification score of the sketch, extracted using the classification Neural Network; the number of strokes performed by the player; and the sketch length, defined as the total number of line segments that defines the sketch. We consider the rest of the available information (country, timestamp...) less relevant for the study performed in this paper.

For the classification Neural Network, a Deep Convolutional Recurrent Neural Network with Long Short Time memory (LSTM) layers, explained in Section \ref{NN}, is implemented. This NN is trained using the Quick, Draw! Dataset. After training, each of the sketches is assigned a score. This score is the ouput assigned by the classification Neural Network to the sketch category, given the sketch as an input to the network. The rest of the parameters can be directly obtained from the dataset.

The remainder of this paper is organised as follows. First, a study of related works using the Quick, Draw! Dataset is presented in Section \ref{ref}. Section \ref{dataset} depicts a more detailed introduction to the dataset. Section \ref{NN} introduces the architecture of the Classification Neural Network implemented. The setup used for the experiments is explained in Section \ref{experiments_setup}. To finish, the final section of results in Section \ref{Results} and conclusions in Section \ref{conclusions}.

\begin{figure}[htpb]
  \centering
  \includegraphics[width=0.15\textwidth]{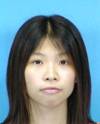}
  \includegraphics[width=0.15\textwidth]{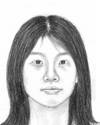}
  \includegraphics[width=0.15\textwidth]{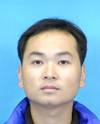}
  \includegraphics[width=0.15\textwidth]{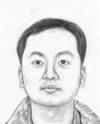}
  \includegraphics[width=0.15\textwidth]{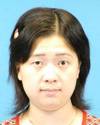}
  \includegraphics[width=0.15\textwidth]{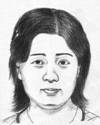}
  \\
  
  \includegraphics[width=0.15\textwidth]{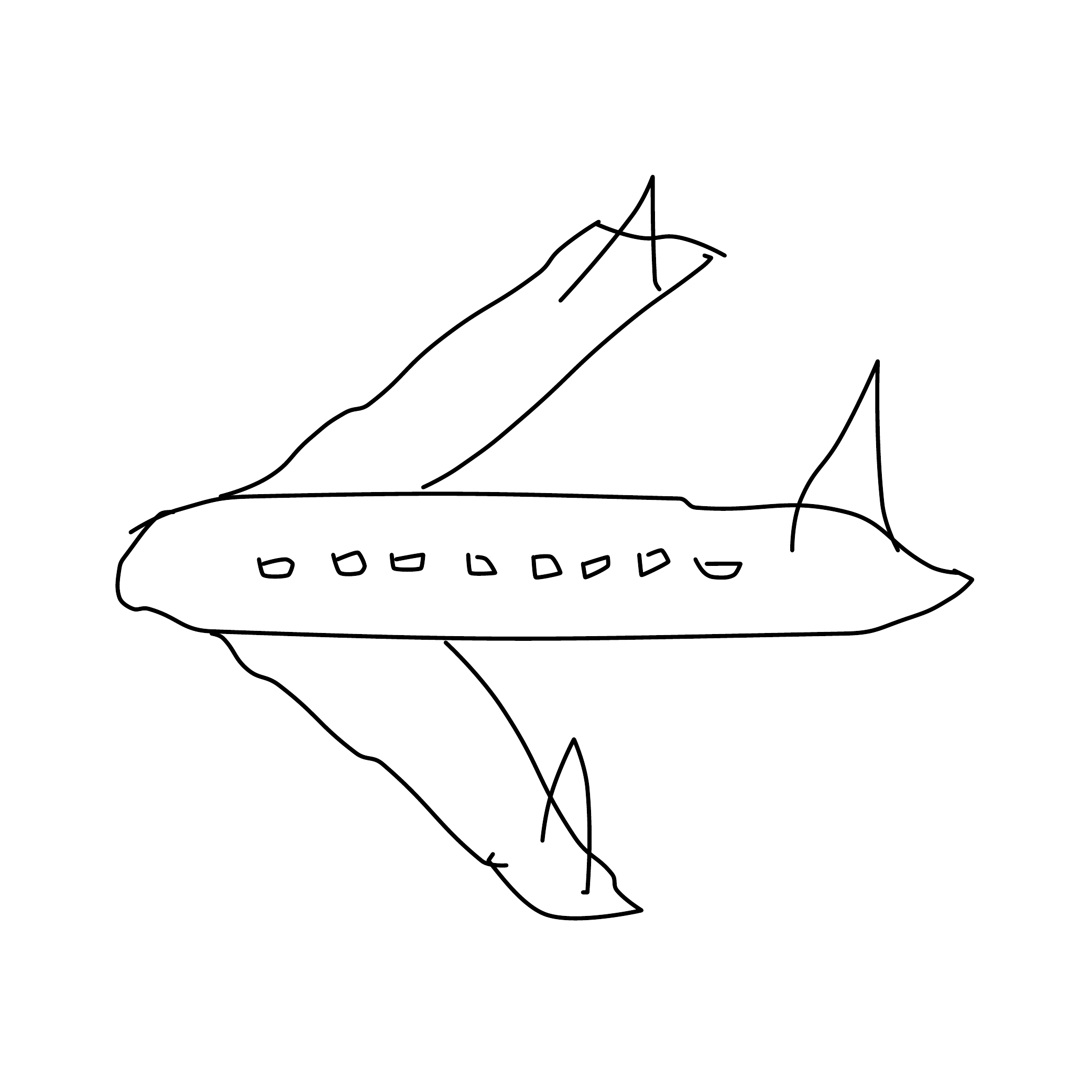}
  \includegraphics[width=0.15\textwidth]{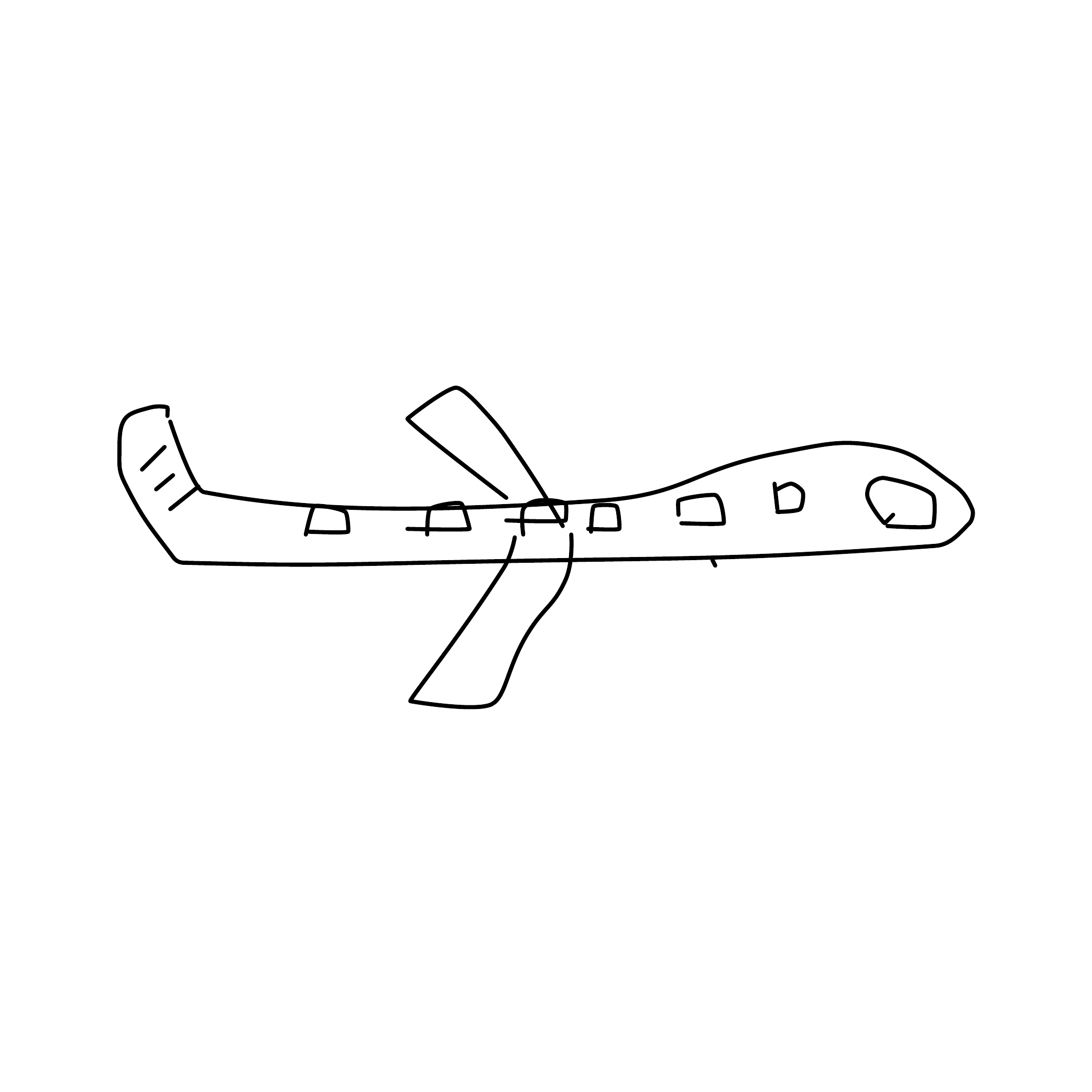}
  \includegraphics[width=0.15\textwidth]{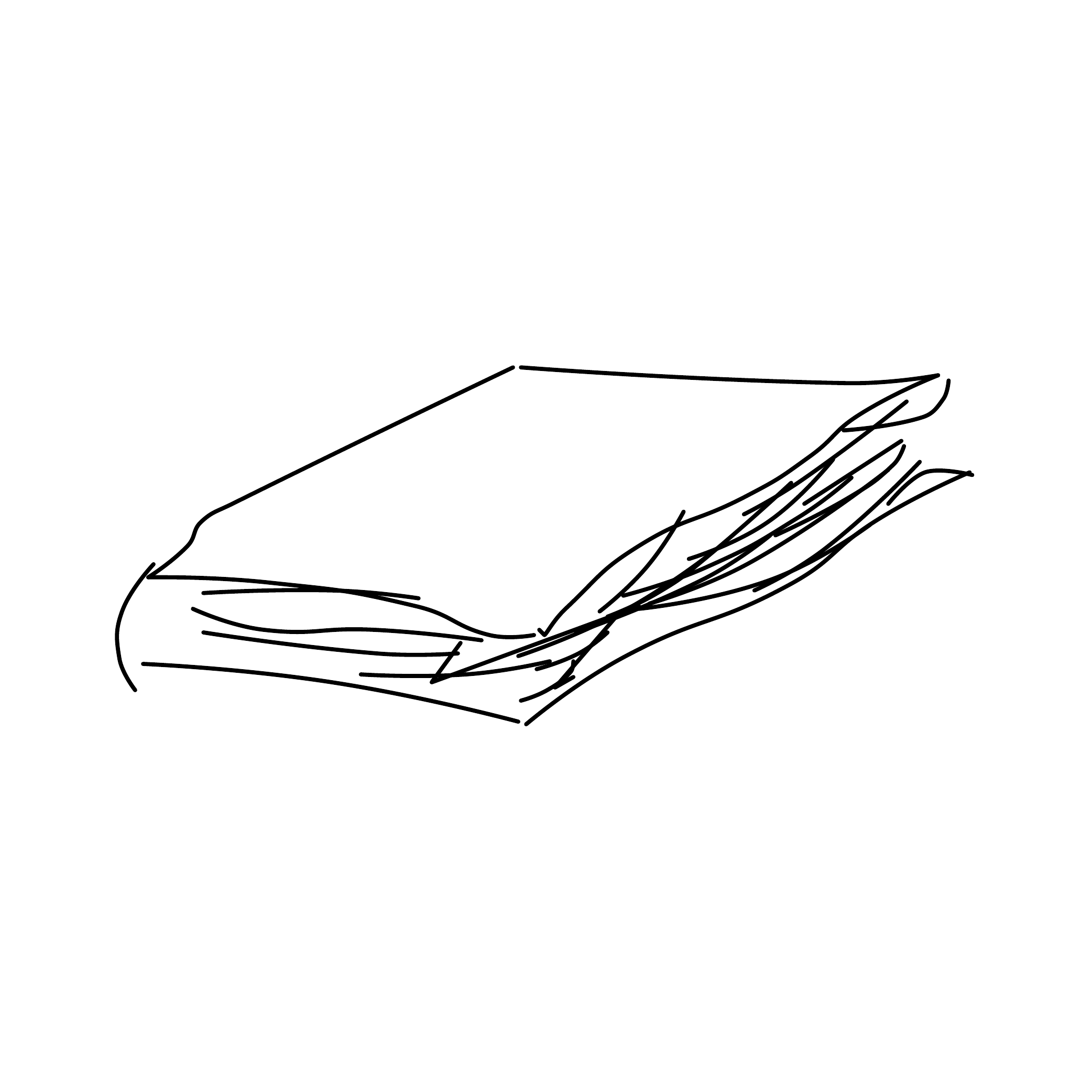}
  \includegraphics[width=0.15\textwidth]{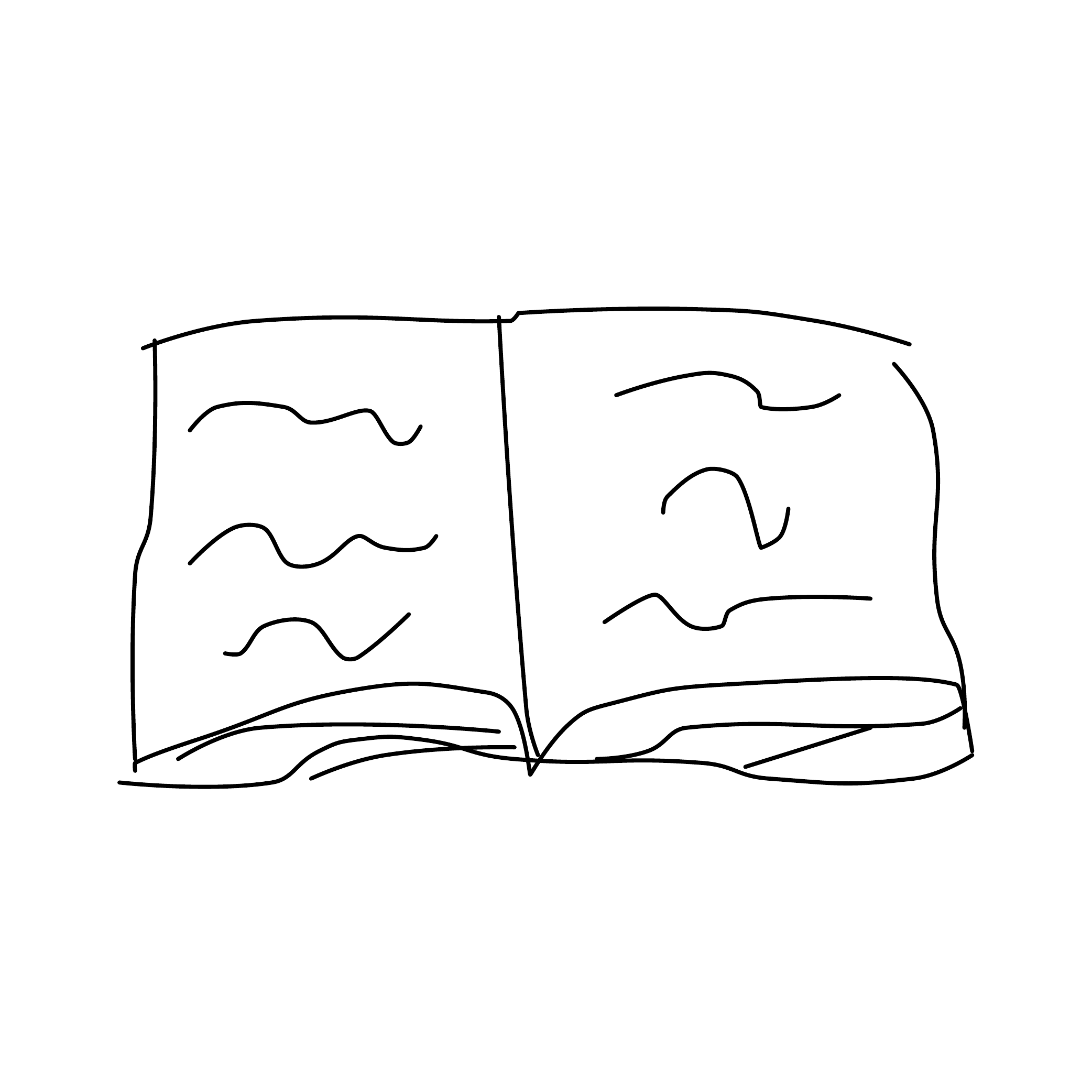}
  \includegraphics[width=0.15\textwidth]{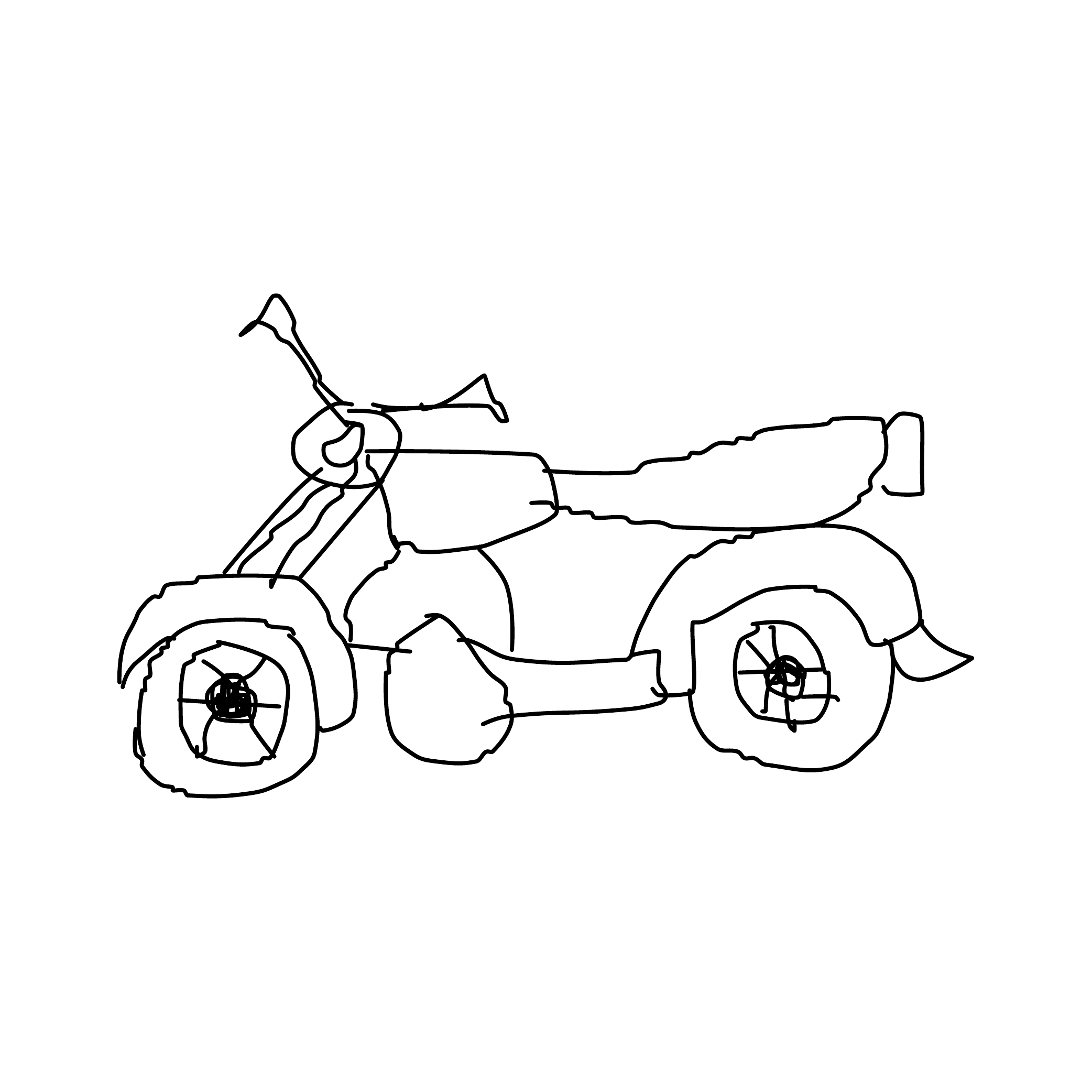}
  \includegraphics[width=0.15\textwidth]{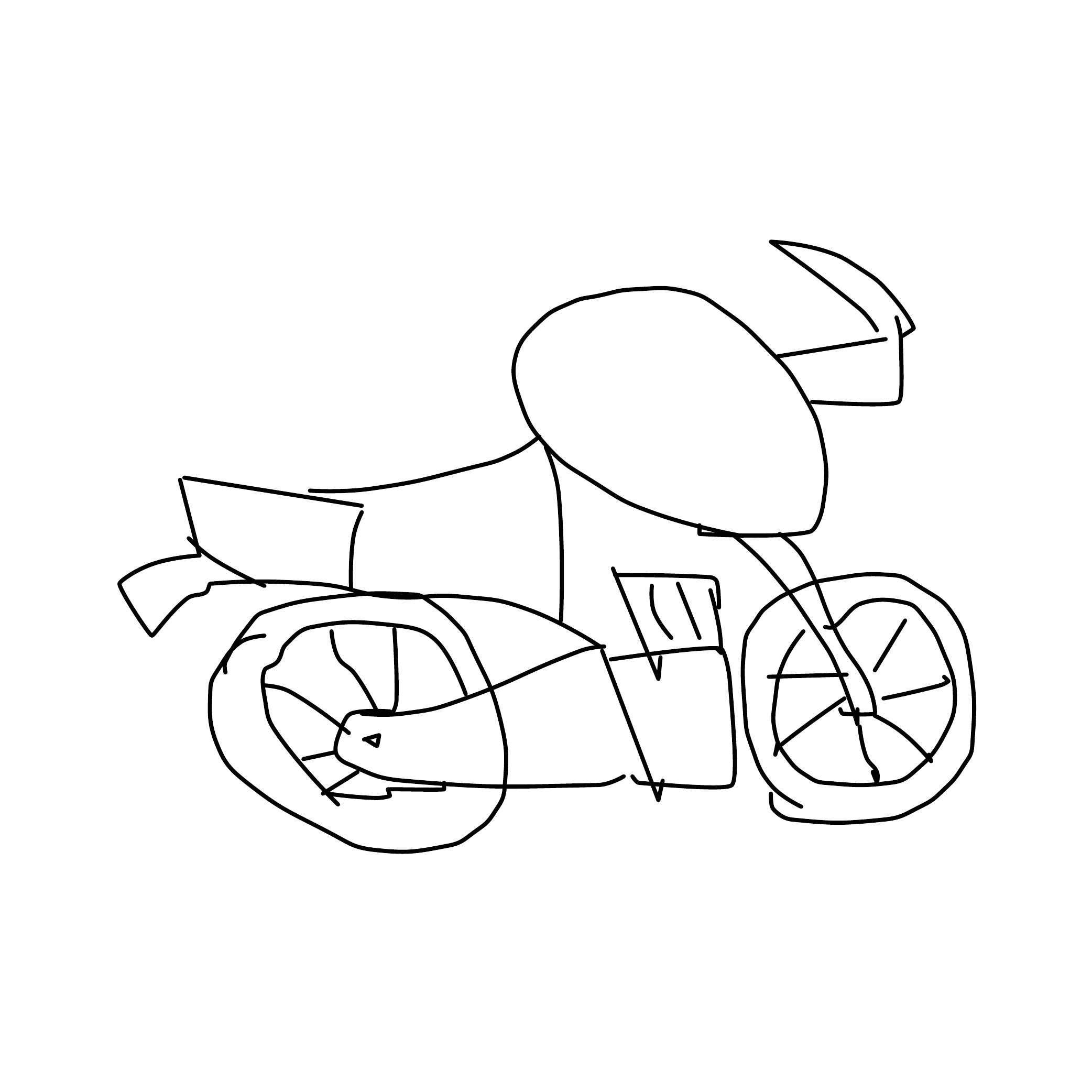}
  \\
  
  \includegraphics[width=0.15\textwidth]{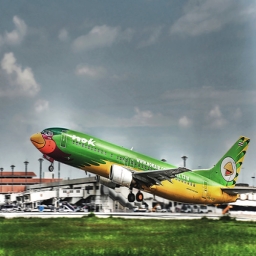}
  \includegraphics[width=0.15\textwidth]{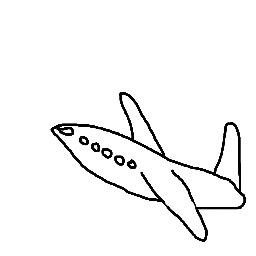}
  \includegraphics[width=0.15\textwidth]{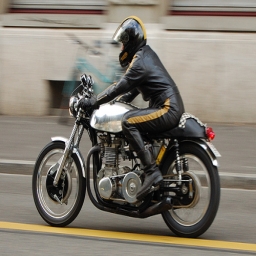}
  \includegraphics[width=0.15\textwidth]{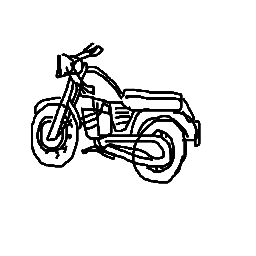}
  \includegraphics[width=0.15\textwidth]{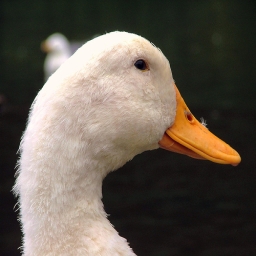}
  \includegraphics[width=0.15\textwidth]{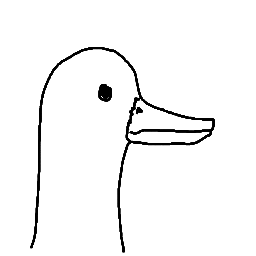}
  \\

  \includegraphics[width=0.15\textwidth]{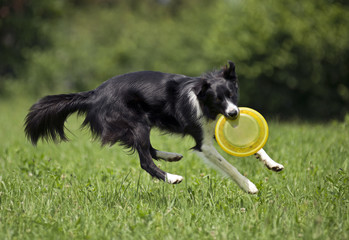}
  \includegraphics[width=0.15\textwidth]{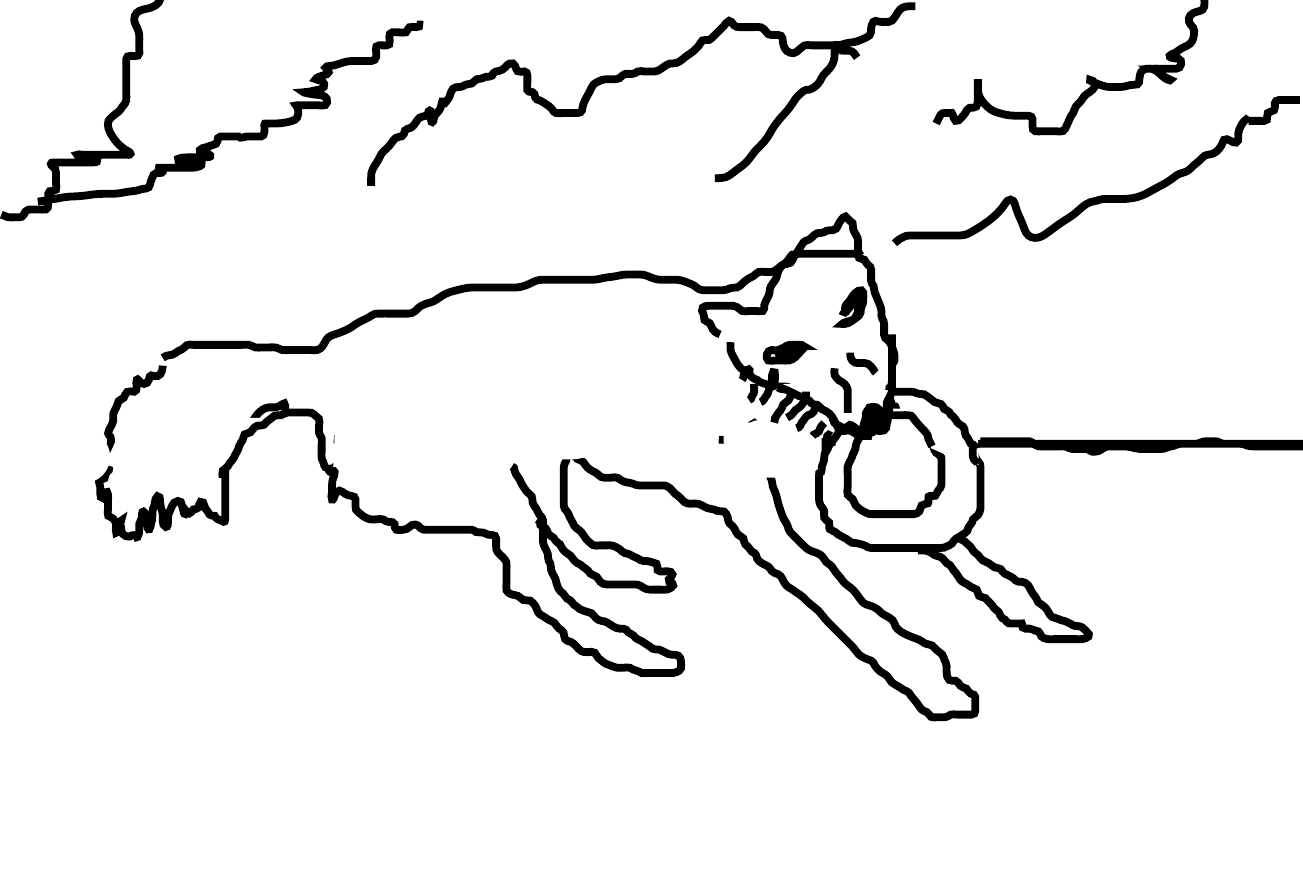}
  \includegraphics[width=0.15\textwidth]{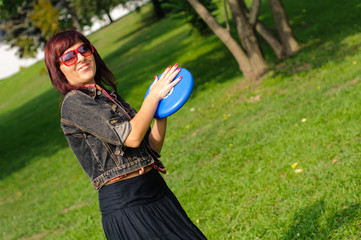}
  \includegraphics[width=0.15\textwidth]{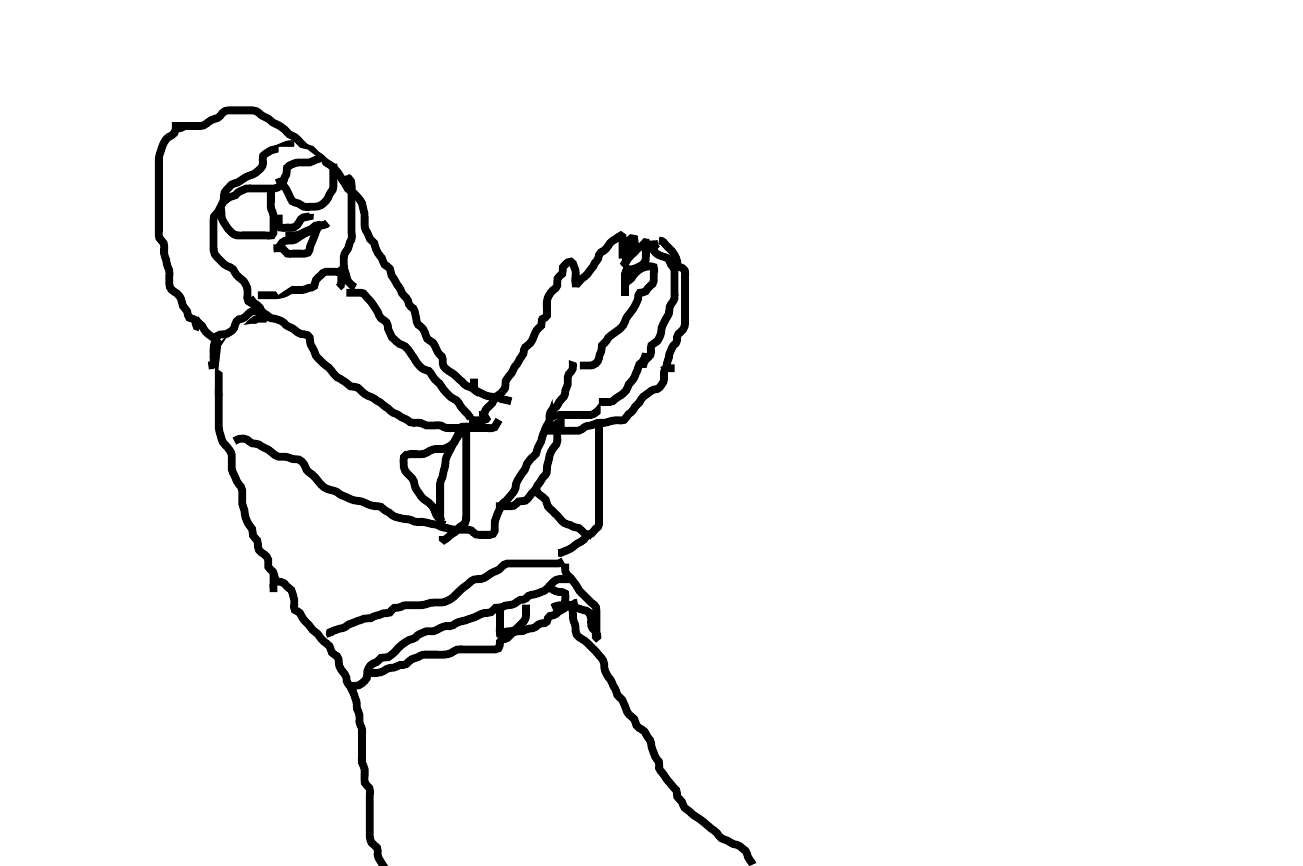}
  \includegraphics[width=0.15\textwidth]{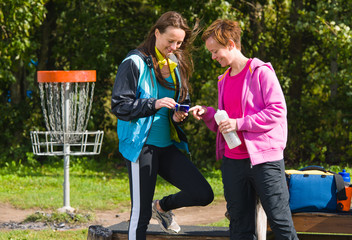}
  \includegraphics[width=0.15\textwidth]{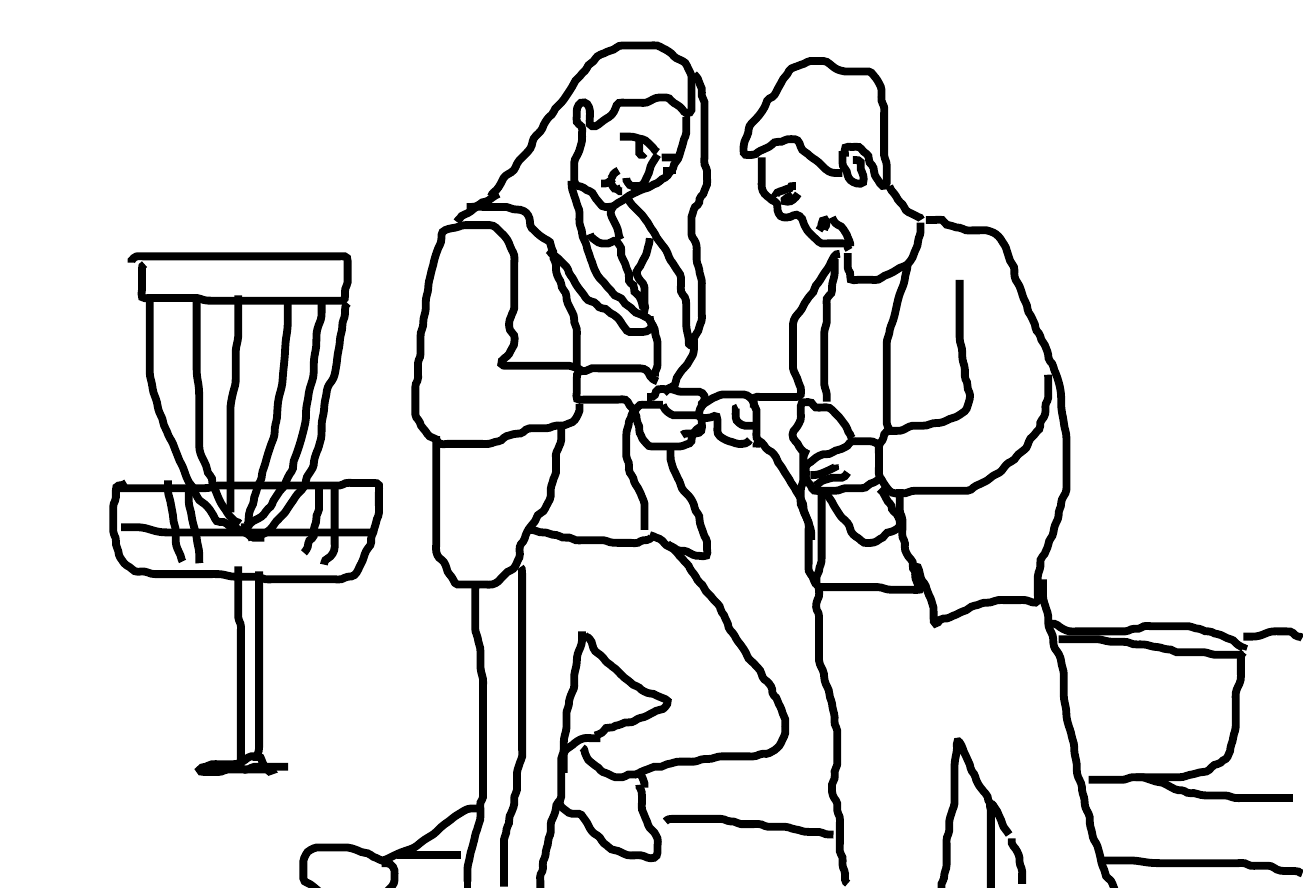}
  \\
  
  \caption{Examples extracted from alternative datasets to the Quick, Draw! Dataset. Each row depicts random examples extracted from a different dataset. From top to bottom: CUHK Face Sketch Database \cite{XiaogangWang2008}; Eitz dataset of human sketches \cite{Eitz2012}; the Sketchy Database \cite{Sangkloy2016} and Li countour dataset \cite{Li2019}. }
  \label{fig:soa}
\end{figure}

\section{State of the Art: Sketch Datasets}
\label{ref}
In addition to the Quick, Draw! Dataset, there are a number of alternatives where the authors created sketch datasets in order to train their machine learning models. The CUHK Face Sketch Database contains a total of 606 sketches of human faces, each with their corresponding photo \cite{XiaogangWang2008}. In the original paper, the authors proposed a Markov Random Field model to synthesize photos to sketches and vice versa. Eitz et al. proposed a dataset containing 20 thousand sketches divided in 250 categories \cite{Eitz2012}. Two different machine learning methods, Nearest-Neighbor and Support Vector Machines, were used for the classification of the sketch dataset. The results of these methods were compared with the ones obtained with human classification. The Sketchy database \cite{Sangkloy2016} presented by Sangkloy et al. is a collection of more than 70 thousand sketches, each paired with a photo, divided in 125 categories. Here, the authors used Convolutional Neural Networks to map the photo and sketches to the same feature space. Recently, Li et al. presented the contour drawing dataset \cite{Li2019}. The countour drawing dataset contain 5000 sketches with 1000 paired photos that can contain multiple objects. The original application of this dataset was to train a deep learning method for sketch generation. All of these datasets are open and can be downloaded in the web page of each project.

\section{Quick, Draw! Dataset}
\label{dataset}
The Quick, Draw! Dataset contains a total of 50 million sketches divided in 345 categories. This dataset is the result of extracting the sketches provided by the players of the online game Quick, draw!.
Compared to state of the art alternatives, it is much larger and may contain much richer information.
As such, the quality of the sketches may experiment important variations, which may be positive for training and posterior recognition.

Sketches are stored in the ``.ndjson'' format\footnote{http://ndjson.org/}. Each sketch is a structure divided in different fields containing the following information: a unique id; the category where the sketch belongs; a recognition flag indicating if the sketch was recognized by the classification network; the timestamp of the sketch; the country of the player; and an array with the drawing information. This drawing information consists of $N$ time series, where $N$ is the number of complete strokes. Each complete stroke is encoded as a $3 \cdot T$ matrix, where the first two rows correspond to the ``x'' and ``y'' position of the pencil, the third one ``t'' with the timestamp, and ``T'' is the number of positions contained within the stroke.

As an alternative to the raw dataset, a simplified dataset is proposed by the Quick, Draw! Dataset authors. In this dataset, the authors introduce the following simplifications:

\begin{itemize}
	\item Align the sketch to the top-left corner, and assign a zero value to that position.
	\item Scale the sketch to a 255x255 canvas.
	\item Re-sample the sketch to have a 1 pixel spacing within strokes.
	\item Use the Ramer–Douglas–Peucker algorithm with an epsilon value of 2.0 to simplify the strokes.
	\item Delete the timestamp information.
\end{itemize}

This simplified dataset is available in three different formats: ``.ndjson'', ``.bin'' and ``.npy''. In this paper the ``.ndjson'' version of the simplified dataset is used as the input to the classification Neural Network. The ``.npy'' version is used for the visualization of sketches. 

\section{Neural Network Classification Model}
\label{NN}

The classification Neural Network takes the sketch input as a sequence of pencil positions of ``x'' and ``y'' and infers the category where the sketch belongs. Due to the nature of the dataset, a Recurrent Neural Network (RNN) using LSTM layers was implemented in this paper \footnote{Adapted from: https://tinyurl.com/tensorflow-models-quickdraw}.
The idea is that in addition to the spatial information contained by the pencil position, there is also a latent space containing relevant temporal information for defining the category of the sketch.
This is where RNN excels. 

\begin{figure}[htpb]
  \centering
  \includegraphics[width=1\textwidth]{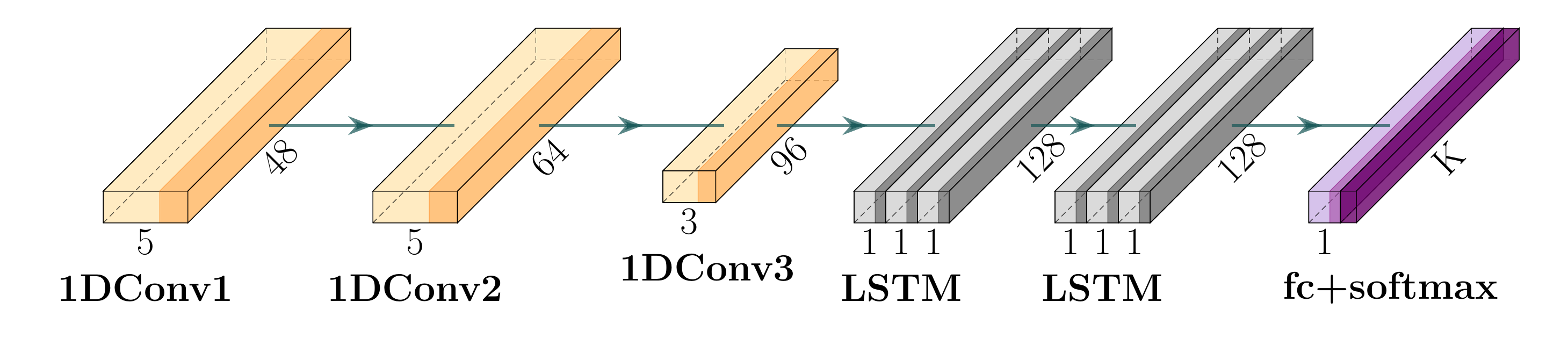}

  \caption{Neural Network classification model.}
  \label{fig:nn-architecture}
\end{figure}

The classification Neural Network model used in this paper is depicted in Fig.~\ref{fig:nn-architecture}. The input of the NN is first passed through three 1D convolution layers. These layers relate the information of the input that is temporally close. The ouptut of the convolution stage is passed to two LSTM blocks using a $tanh$ activation function. The LSTM blocks are composed by three LSTM layers each. The final layer is a Fully Connected (FC) layer with a softmax activation, being the outputs the score assigned for each of the $K$ categories, where $K$ is the number of training categories. For the training of the Neural Network, a total of 10000 sketches were used for the training stage with 1000 sketches for the validation stage for each category.

\section{Experimental Setup}
\label{experiments_setup}

The goal of this paper is to provide a statistical analysis of the Quick, Draw! Dataset. Three parameters will be extracted to perform this analysis: the score obtained with the classification Neural Network proposed in section \ref{NN}; the total sketch length; and the total number of complete strokes defined in each sketch. The Quick, Draw! Dataset is composed by 345 categories, each of these containing hundreds of thousands of sketches. Training a NN that is able to classify all of the categories is computationally expensive. In this paper we choose three of these categories to train the classification Neural Network. The goal is to obtain representative statistical results that can give the reader a first impression of the spectrum of sketches contained in this dataset. 

The three categories studied in this paper are: \emph{mountain}, \emph{book} and \emph{whale}. 
These three categories go from the simplicity of a mountain to the complexity of a whale, being the book the average level sketch example. The three parameters used in this paper were chosen to extract the most relevant information from the sketches. The classification score provides a value of the quality of the sketch. The sketch length and number of strokes provides information about the complexity of the sketches. Together, these three parameters provide information about the quality and complexity of the sketches.

In the following section, the results of these experiments 
are depicted. All the sketches contained in each of the categories studied in this paper at the time of its writing were used to extract these results.

\section{Experimental Results}
\label{Results}

At the time of writing this paper, a total of $364.409$ sketches were contained over the three categories studied. The distribution of these sketches between the categories was the following: $128.541$ \emph{mountain} sketches; $119.365$ \emph{book} sketches and $116.503$ \emph{whale} sketches. All of the sketches were used in order to obtain the results presented in this paper.

\begin{table*}[thpb]
\caption{Experimental results for the statistical analysis of the Quick, Draw! Dataset as a function of the three categories studied in this paper, where $\mu$ is the arithmetic mean, and $\sigma$ is the standard deviation.}
\label{tab:res}
\centering
\begin{tabularx}{\textwidth}{|c *{7}{|Y}|}
	\cline{2-7}
	 \multicolumn{1}{c|}{} & \multicolumn{2}{c|}{Classification score} & \multicolumn{2}{c|}{Number of strokes} & \multicolumn{2}{c|}{Sketch length} \\ \hline
	 Category & $\mu$ & $\sigma$ & $\mu$ & $\sigma$ & $\mu$ & $\sigma$  \\ \hline
	 \emph{mountain} & 5.16 & 2.13 & 2.09 & 2.15 & 18.04 & 15.08  \\ \hline
	 \emph{book}  & 2.96 & 2.67 & 7.21 & 3.78 & 42.50 & 17.17 \\ \hline
	 \emph{whale}  & 4.44 & 2.49 & 5.55 & 3.45 & 45.38 & 17.68 \\ \hline
	\end{tabularx}
\end{table*}

\begin{figure*}[htpb] 
  \centering
  \includegraphics[width=0.8\textwidth]{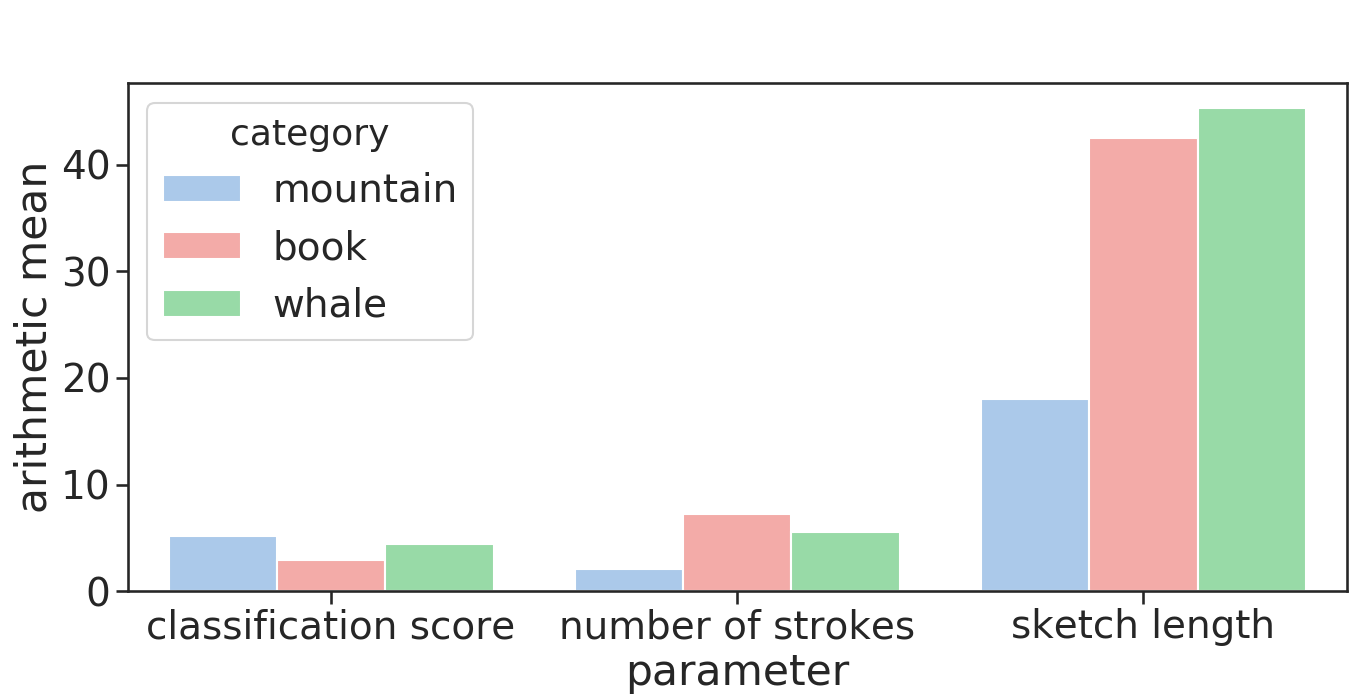}
  \caption{Comparison between the results obtained with the three categories studied in this paper.}
  \label{fig:barplot}
\end{figure*}

Table \ref{tab:res} depicts the arithmetic mean and the standard deviation of the distribution of sketches as a function of the three parameters studied in this paper. From these results, it can be concluded that the \emph{book} is the category with a lower average classification score. This is interesting, since the \emph{whale} category was initially a more complex one, while it has a higher average classification score. There are two possibilities that can explain this behavior: either people perform poorly at drawing books or the classification network performs poorly at book classification. If we take a look at the second and third parameters depicted on the table, in average, \emph{book} sketches are composed by a 30\% more strokes than \emph{whale} ones do with a similar length in the total drawing. Although the concept of drawing a book looks easier than that of drawing a whale, people tend to draw more complex books than whales. This can lead to a loss in the quality of the drawings, or a loss in the classification network performance. 
Fig. \ref{fig:barplot} provides a graphical comparison between the results obtained for each of the three parameters studied as a function of the category.

\begin{table*}[thpb]
\caption{Number of sketches as a function of its z-score divided by categories. Only the classification score is taken into account to compute this z-score.}
\label{tab:z-score}
\centering
\begin{tabularx}{\textwidth}{|c *{11}{|Y}|}
	 \cline{2-11} 
	 \multicolumn{1}{c|}{} & \multicolumn{10}{c|}{z-score (classification score) } \\ \hline
	 category & $-\sigma$ & $\sigma$ & $-2\sigma$ & $2\sigma$ & $-3\sigma$ & $3\sigma$ & $-4\sigma$ & $4\sigma$ & $-5\sigma$ & $5\sigma$  \\ \hline
	 \emph{mountain} & 3042 & 2034 & 1725 & 273 & 498 & 12  & 60 & 7 & 3 & 4\\ \hline
	 \emph{book}  & 10245 & 2687 & 1383 & 126 & 267 & 16 & 69 & 2 & 21 & 0 \\ \hline
	 \emph{whale}  & 2064 & 5165 & 78 & 279 & 15 & 51 & 9 & 11 & 0 & 7 \\ \hline
	\end{tabularx}
\end{table*}

\begin{figure*}[h!]
  \includegraphics[width=0.5\textwidth]{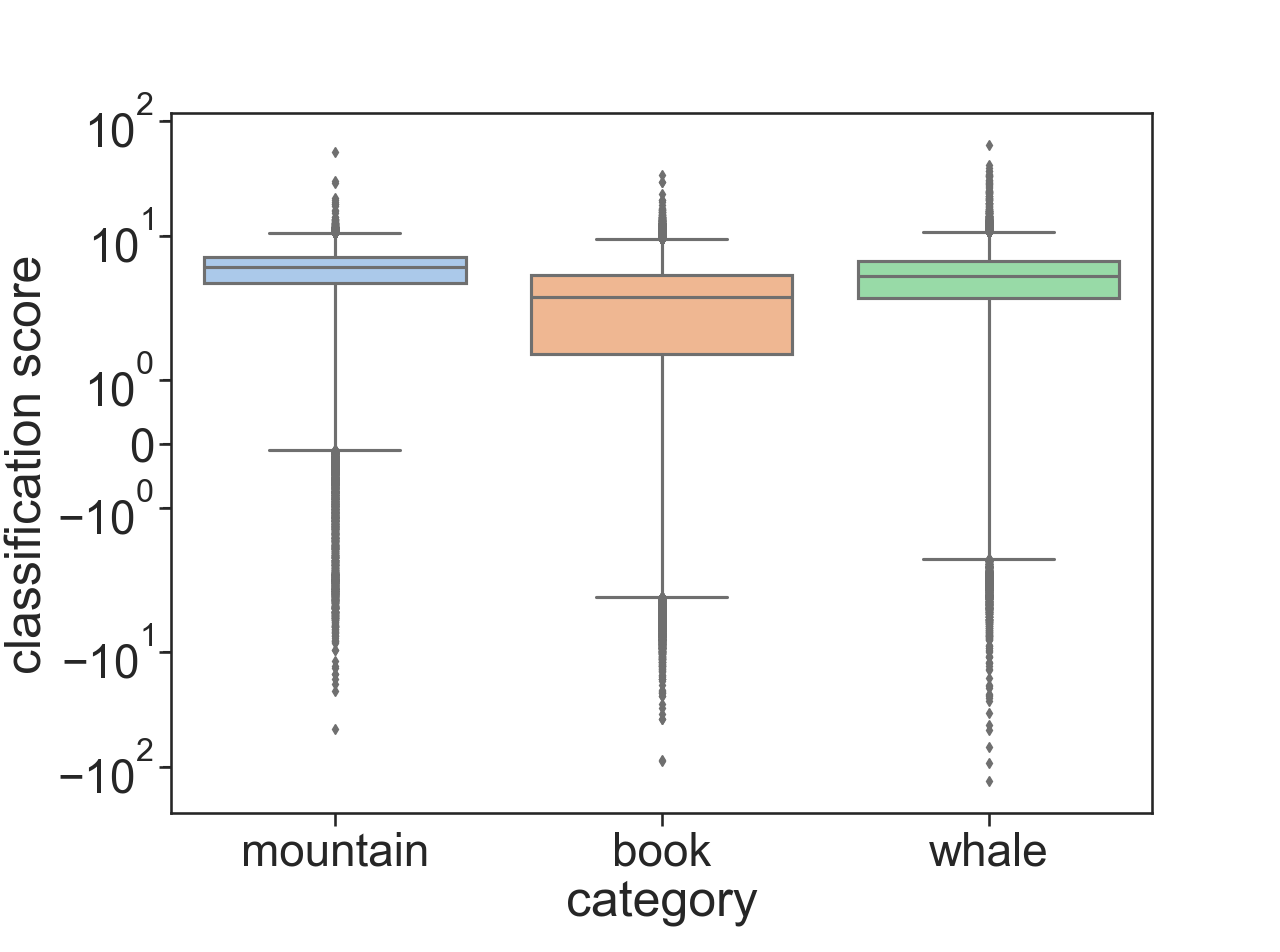}
  \includegraphics[width=0.5\textwidth]{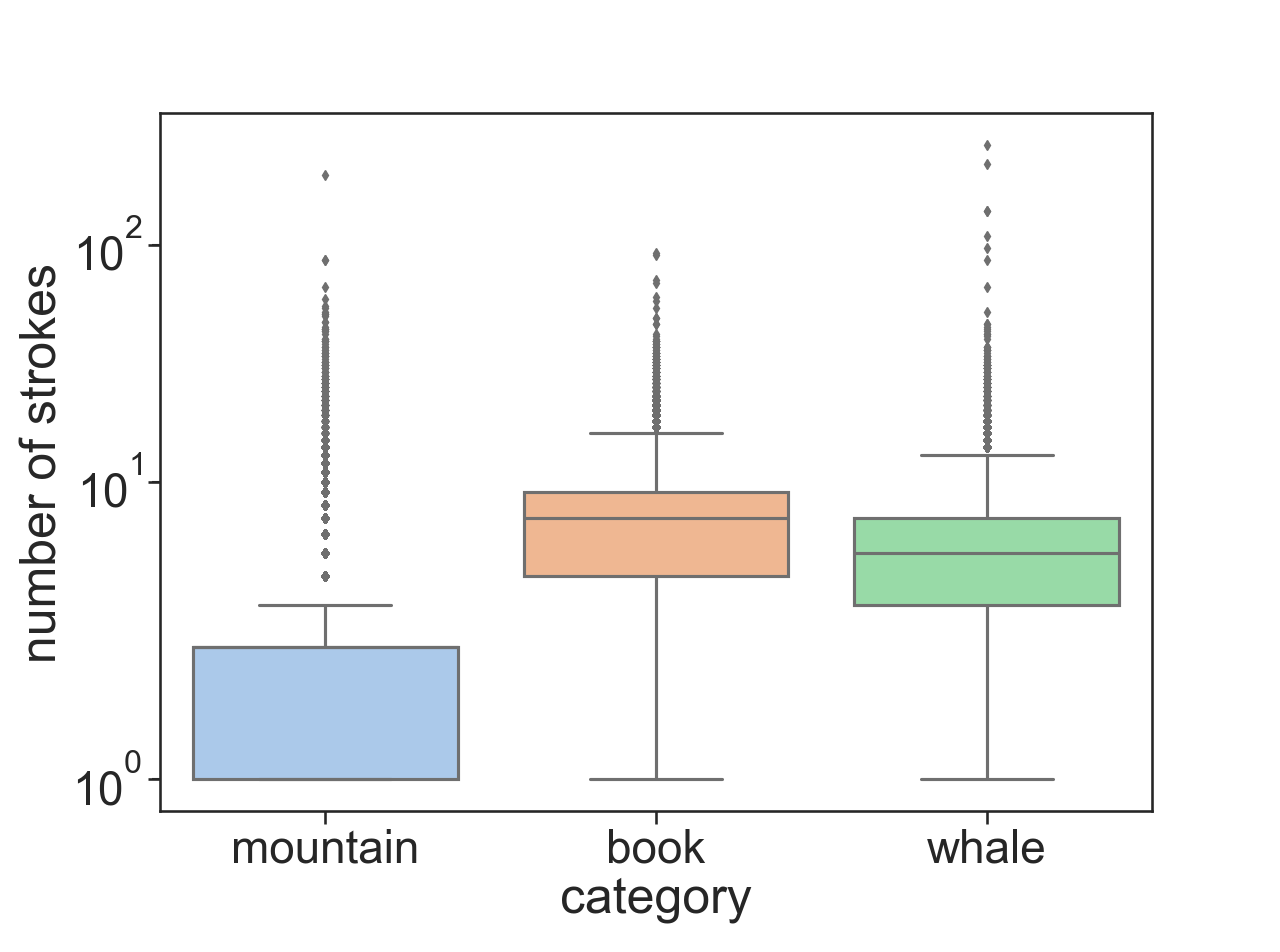}
  \
  \begin{center}
  \includegraphics[width=0.5\textwidth]{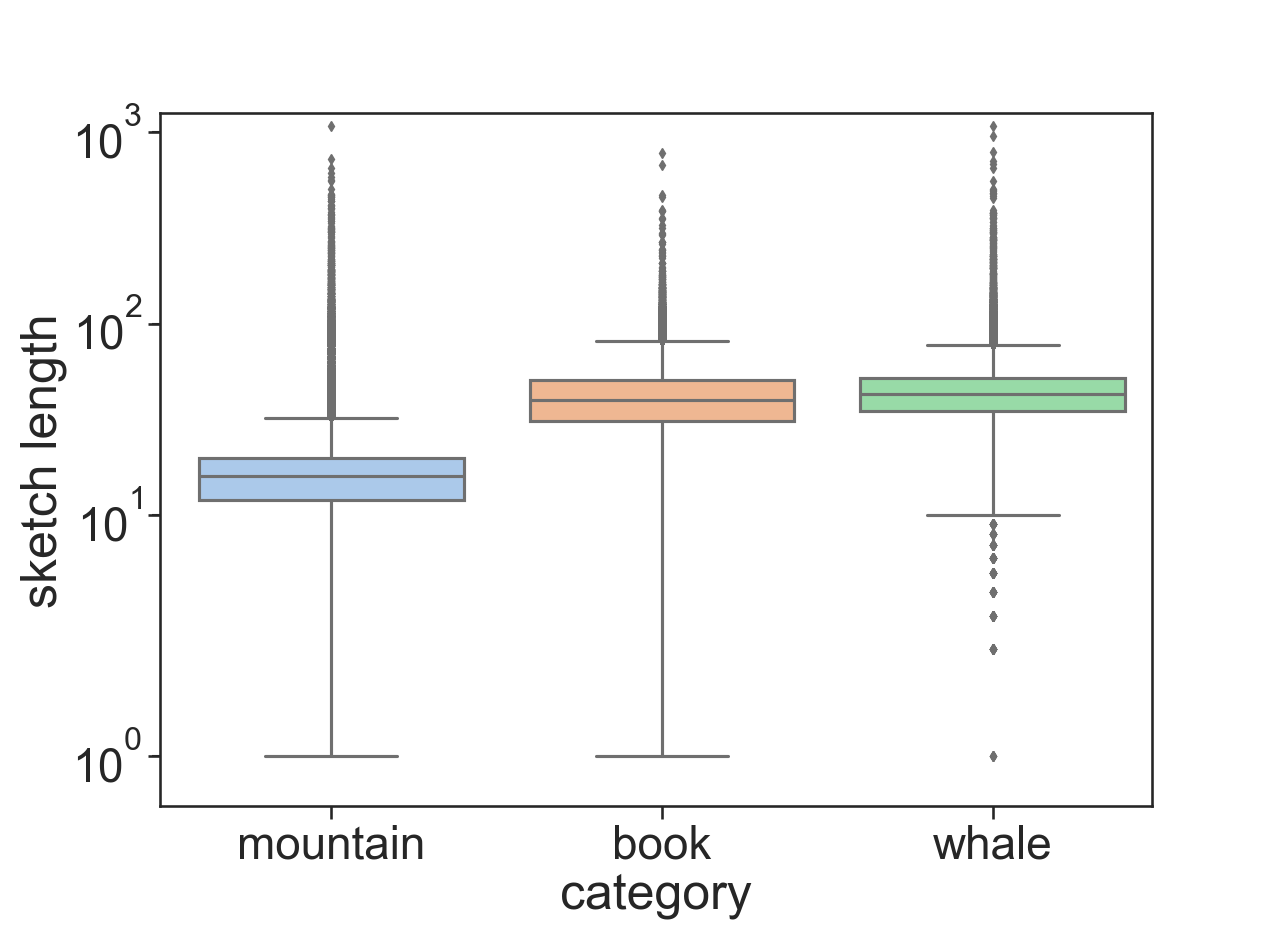}
  \end{center}
  \caption{Comparison between the results obtained with the three categories studied in this paper.}
  \label{fig:boxplot}
\end{figure*}

\begin{figure*}[htpb] 
  \centering
  \includegraphics[width=0.8\textwidth]{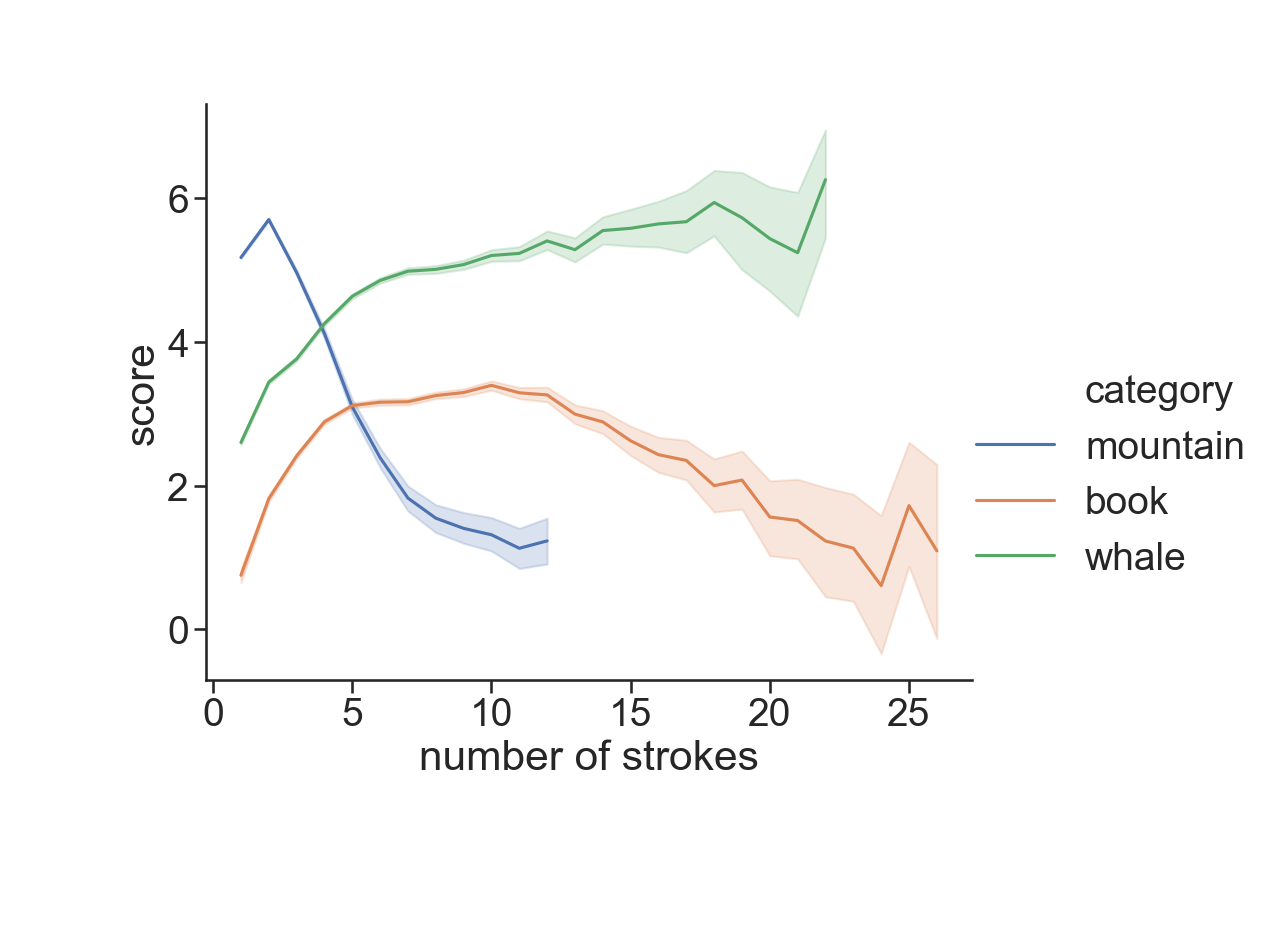} \\
  \hspace*{0.2cm}
  \includegraphics[width=0.8\textwidth]{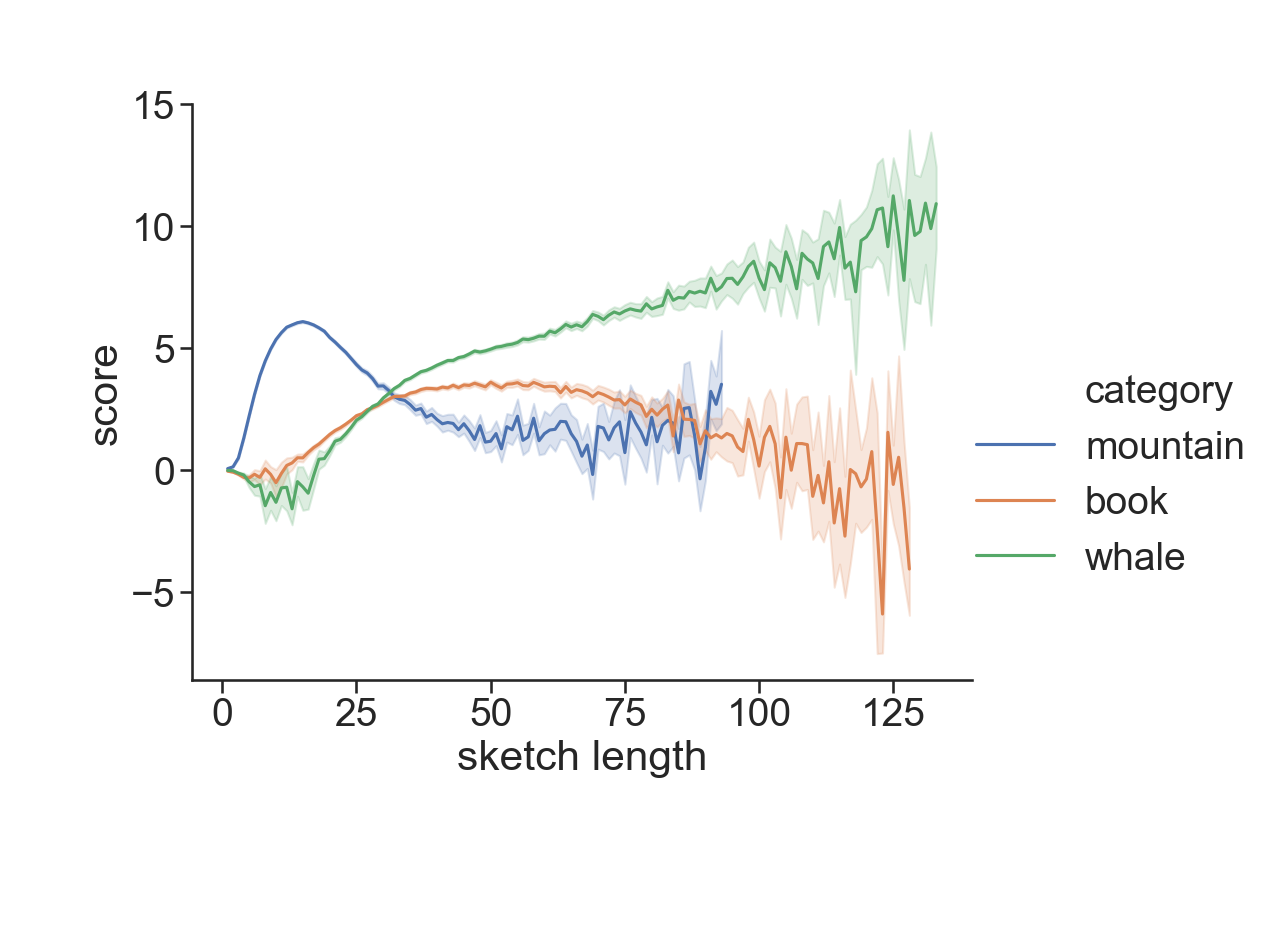}
  \caption{Classification score as a function of the sketch length and number of strokes for the three categories presented in this paper. Sketches with a z-score bigger than 5 were not depicted in the graph for clearness reasons.}
  \label{fig:relation}
\end{figure*}

Table \ref{tab:z-score} contains the number of sketches as a function of its z-score. This z-score gives an impression of the number of outliers in the dataset. In the case of the \emph{mountain} and \emph{book} sketches the majority of outliers are in the negative region. In these two categories the population of outliers are mainly composed by people performing poor sketches. In the case of the \emph{whale} category, the population of outliers is larger in the positive region meaning an outlier population of people performing better sketches. This gives us a first impression of the differences in the complexity of the sketches. In the two first categories, the more outlying sketches are the ones that are poorly drawn. In the case of the \emph{whale} the more outlying sketches are the one that achieve the higher level of classification score. The results in this table also show how the number of outliers is not as high as expected for a dataset of this nature. In the case of the \emph{mountain} category, around 96\% of the population of mountain sketches are contained in the range $(-\sigma, \sigma)$,  89\% for \emph{book} sketches, and 94\% for \emph{whale}.  Here, it should be noted that, as extracted from the Quick, Draw! Dataset repository, the dataset authors claim to have moderated the collection of drawings individually. Fig. \ref{fig:boxplot} depicts various boxplot graphs for a better visualization of the distribution of sketches as a function of the three parameters and the category where they belong.

Fig. \ref{fig:relation} depicts the relation between the classification score and the number of strokes and length of the sketches. The first thing to take into account is that both the sketch length and the number of strokes have a similar effect in the classification score achieved in each category. In the case of the \emph{mountain} category, the peak of classification score is achieved with low values of these two parameters. Increasing the number of strokes or the sketch length comes with a loss in the classification score for this category. More simple sketches are preferred by the neural network for the classification of \emph{mountain} sketches. The case of the \emph{book} category in terms of shape is quite similar. The only difference here is that this peak is achieved with higher levels of strokes and sketch length. As in the \emph{mountain} category, once reached this peak, the classification score of the sketches decrease. In the case of the \emph{whale} category, increasing the complexity of the sketches comes in average with an increase of the classification score obtained by these sketches. In this case, high complexity sketches are, in average, better in quality. This comes with a cost: higher complexity sketches also mean higher levels of variability in the quality of sketches. This effect can be found in the three categories studied; however, it is more relevant in the case of the \emph{whale} category, since the best sketches are the ones with a higher complexity. A similar effect happens for low quality sketches in the case of the \emph{book} and \emph{whale} categories. For these two categories low quality sketches suffer from high variances in quality. This effect does not occur in the case of the \emph{mountain} category.

\section{Conclusions}
\label{conclusions}

The current tendency to increase the number of layers in Deep Neural Networks have produced models with hundreds of layers. These models require very large amounts of data to provide useful results. The Quick, Draw! Dataset, composed by over 50 million sketches, is presented as way to deal with this need of massive data for training. In this paper, a statistical study of three of the categories present in this dataset: \emph{mountain}, \emph{book} and \emph{whale} is presented. The results depict the distribution of all the sketches of these three categories of the dataset as a function of three parameters: classification score, number of strokes, and sketch length. For the experiments, a classification Neural Network using LSTM layers was trained and used for prediction. The results obtained in this paper have depicted some interesting results regarding the nature of the sketches contained in this dataset.

The first relevant conclusion is that the complexity of a sketch is not only limited to the complexity of the category as would be expected. Conceptually, describing a \emph{whale} is more complex than describing a \emph{book}, however the results obtained in this paper show that in average, people drew more complex sketches to define a \emph{book} than to define a \emph{whale}. A similar concept was already introduced by \cite{Sangkloy2016}, where a ``sketchability'' criterion was defined as a subjective measure of how hard it would be to sketch some of the photos. Some interesting studies can be extracted about the differences in sketch complexity between the different categories. 

Another interesting result extracted from the experiments is that the number of outliers presented in the dataset is surprisingly low for a dataset of this nature. The Quick, Draw! Dataset's source is the open online game Quick, Draw!, where hundred of thousands of players around the world have accessed this game and sketched. With this premise, one would expect for this dataset to have high variances in quality between the sketches in a category, see Fig. \ref{fig:3class} for reference. The results in the experiments depict that more than 90\% of the sketches in average are contained in the range $(-\sigma, \sigma)$ for the categories presented. In addition to this, the number of outliers outside the range $(-3\sigma, 3\sigma)$ is lower than the 0.05\% of the total population in each category studied. In the repository of this dataset, the authors claim to have performed a individual analysis of the sketches. This could explain the results obtained. However, there is no information about how this analysis was performed, and as Fig. \ref{fig:3class} depicts it was not perfect. This should be taken in account before considering this dataset a raw dataset or considering all the sketches in this dataset raw examples of each category.  

\section{Applications and Future Work}
The Quick, Draw! Dataset may fit in two kinds of applications: applications where very large amounts of data are required (e.g. training Deep Neural Networks) and applications where the trajectory performed during the sketch, rather than the image, is required. Certain precautions must be taken into account during the interpretation of its contents, as stated in the dataset repository\footnote{https://github.com/googlecreativelab/quickdraw-dataset} and depicted in the results of this paper, this is not a raw dataset. Any conclusions extracted from the analysis of this dataset could be altered by possible exclusions resulting of the reviewing process. 
Future works using this dataset involves a more complete analysis using more categories as input, and the study of more parameters that can be interesting for machine learning techniques.

For downloading, obtaining more information or checking interesting projects involving this dataset, refer to the online repository referenced in this paper.

\section{Acknowledgment}
The research leading to these results has received funding from RoboCity2030-DIH-CM, Madrid Robotics Digital Innovation Hub, S2018/NMT-4331, funded by “Programas de Actividades I+D en la Comunidad de Madrid” and cofunded by Structural Funds of the EU.

\end{document}